\documentclass[acmsmall]{acmart}

\AtBeginDocument{%
  }

\setcopyright{cc}
\setcctype{by-nc-sa}
\acmJournal{THRI}
\acmYear{2025} \acmVolume{14} \acmNumber{3} \acmArticle{46} \acmMonth{4} \acmDOI{10.1145/3719020}

\usepackage{multicol}
\usepackage{graphicx} 
\usepackage[shortlabels]{enumitem}
\usepackage[export]{adjustbox}
\usepackage{booktabs,colortbl,tabularx}
\usepackage{multirow}
\usepackage{amsmath, amsfonts}
\usepackage{siunitx}
\usepackage{caption}
\usepackage{xcolor}

\newcommand{\track}[1]{\textcolor{black}{#1}}

\newcommand{\thri}[1]{\textcolor{black}{#1}}

\newcommand{\thrirevision}[1]{\textcolor{black}{#1}}

\colorlet{BLACK}{black}

\begin{document}

\title{Predicting Human \thrirevision{Perceptions} of Robot Performance During Navigation Tasks}

\author{Qiping Zhang}
\authornote{Equally contributing authors}
\affiliation{%
  \institution{Yale University}
  \city{New Haven, CT}
  \country{USA}}
\email{qiping.zhang@yale.edu}

\author{Nathan Tsoi}
\authornotemark[1]
\affiliation{%
  \institution{Yale University}
  \city{New Haven, CT}
  \country{USA}}
\email{nathan.tsoi@yale.edu}

\author{Mofeed Nagib}
\affiliation{%
  \institution{Yale University}
  \city{New Haven, CT}
  \country{USA}}
\email{mofeed.nagib@yale.edu}

\author{Booyeon Choi}
\affiliation{%
  \institution{Yale University}
  \city{New Haven, CT}
  \country{USA}}
\email{brian.choi@yale.edu}

\author{Jie Tan}
\affiliation{%
  \institution{Google DeepMind, Google Inc}
  \city{Mountain View, CA}
  \country{USA}}
\email{jietan@google.com}

\author{Hao-Tien Lewis Chiang}
\affiliation{%
  \institution{Google DeepMind, Google Inc}
  \city{Mountain View, CA}
  \country{USA}}
\email{lewispro@google.com}

\author{Marynel V{\'a}zquez}
\affiliation{%
  \institution{Yale University}
  \city{New Haven, CT}
  \country{USA}}
\email{marynel.vazquez@yale.edu}

\renewcommand{\shortauthors}{Q. Zhang et al.}


%

\begin{abstract}
\thrirevision{Understanding human perceptions of robot performance is crucial for designing socially intelligent robots that can adapt to human expectations. Current approaches often rely on surveys, which can disrupt ongoing human-robot interactions. As an alternative, we explore predicting people's perceptions of robot performance using non-verbal behavioral cues and machine learning techniques.}
We contribute the SEAN TOGETHER Dataset consisting of observations of an interaction between a person and a mobile robot in Virtual Reality, together with \thrirevision{perceptions} of robot performance provided by users on a 5-point scale. We then analyze how well humans and supervised learning techniques can predict perceived robot performance based on different observation types (like facial expression and spatial behavior features). Our results suggest that facial expressions alone provide useful information; but in the navigation scenarios that we considered, reasoning about spatial features in context is critical for the prediction task. Also, supervised learning techniques outperformed humans' predictions in most cases. Further, when predicting robot performance as a binary classification task on unseen users' data, the $F_1$-Score of machine learning models more than doubled that of predictions on a 5-point scale. This suggested good generalization capabilities, particularly in identifying performance directionality over exact ratings. Based on these findings, we conducted a real-world demonstration where a mobile robot uses a machine learning model to predict how a human who follows it perceives it. Finally, we discuss the implications of our results for implementing \thrirevision{these} supervised learning models in real-world navigation. \thrirevision{Our work paves the path to automatically enhancing robot behavior based on observations of users and inferences about their perceptions of a robot.
}
\end{abstract}


\maketitle

\section{Introduction}
\label{sec:introduction}
As a scalable alternative to measuring subjective \thrirevision{perceptions} of robot performance through surveys, recent work in Human-Robot Interaction (HRI) has explored using \textit{implicit} human feedback to predict these \thrirevision{perceptions} \cite{aronson2018gaze,cui2021empathic,stiber2022modeling,zhang2023self}. These are communicative signals that are unintentionally exhibited by people 
\cite{knepper2017implicit}. They can be reflected in human actions that change the world's physical state \cite{sadigh2016planning} or can be nonverbal cues, such as facial expressions \cite{cui2021empathic,stiber2022modeling} and gaze \cite{mitsunaga2008adapting,aronson2018gaze}, 
displayed during social interactions. Implicit feedback serves as a burden-free information channel 
that sometimes persists even when people don't intend to communicate  \cite{kendon1988goffman}.

We expand the existing line of research on predicting \thrirevision{perceptions} of robot performance from nonverbal human behavior to dynamic scenarios involving  robot navigation. 
Prior work 
has often considered stationary tasks, like physical assembly at a desk \cite{stiber23hri} or robot photography \cite{zhang2023self}, in laboratory environments. 
We instead explore the potential of using observations of the body motion, gaze, and facial expressions of a person to predict their \thrirevision{perceptions} of a robot's performance while a robot guides them to a destination in a crowded environment. These \thrirevision{perceptions} correspond to subjective opinions of how well a robot is performing the navigation task. Predicting them in crowded navigation scenarios is more challenging than in stationary settings because human nonverbal behavior can be a result of not only robot behavior, but also other interactants in the environment. Further, because of motion, nonverbal responses to the robot may change as a function of the environment. For example, imagine that the person that follows the robot looks downwards. This could reflect paying attention to the robot, or be a result of the person inspecting their nearby physical space, which varies during navigation.

\begin{figure}[t!p]
    \centering
    \includegraphics[width=0.45\linewidth]{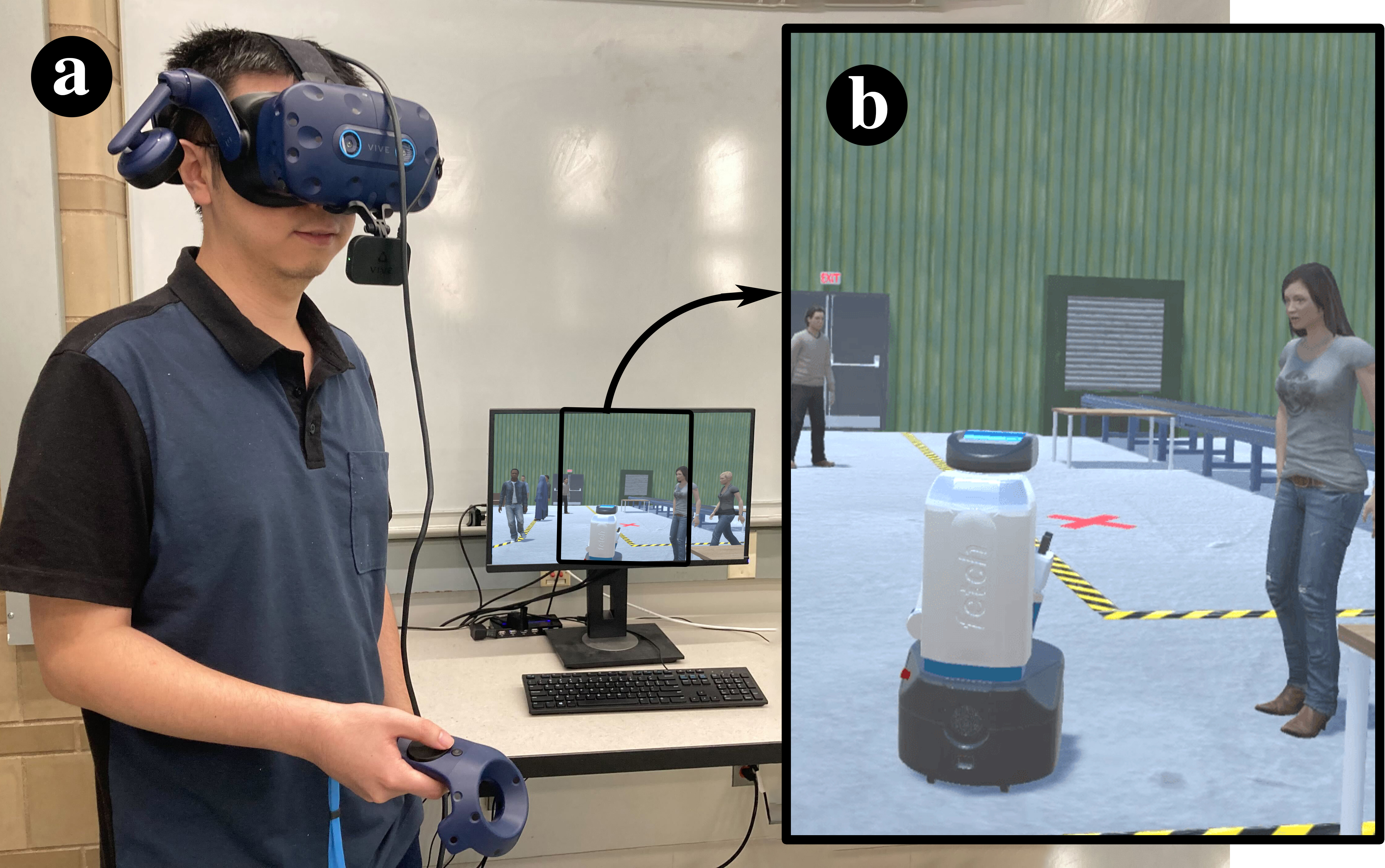}

    \caption{Data collection.  Humans controlled an avatar in the simulation with VR (a) while they were guided by a Fetch robot (b). The screen on the desk shows what the user saw. 
    }
    \label{fig:setup}
\end{figure}
 
To study implicit feedback during navigation tasks, we performed a systematic data collection  using the Social Environment for Autonomous Navigation (SEAN) 2.0~\cite{tsoi2022sean} with Virtual Reality (VR)  \cite{zhang2023sean}.\footnote{Dataset \thrirevision{and source code} available at: \url{https://sean-together.interactive-machines.com/}.}
Humans took part in the simulations through an avatar, which was controlled using a VR headset, as in Fig.~\ref{fig:setup}.
The headset enabled immersion and allowed us to capture implicit feedback features like gaze. Also, it  facilitated querying the human about robot performance as navigation tasks took place. We considered robot performance as a multi-dimensional construct, similar to \cite{zhang2023self}, because humans may care about many aspects of a robot's navigation behavior, as discussed in the social robot navigation literature \cite{gao2022evaluation,mavrogiannis2023core,francis2023principles}. 

Then, we \track{studied} fundamental questions about the value of implicit feedback signals in predicting subjective \thrirevision{perceptions} of robot performance \track{using the VR data}. First, we investigate\track{d} to what extent humans can predict a person's \thrirevision{perceptions} of the robot's performance (along the dimensions of perceived \thrirevision{navigation} competence, \thrirevision{surprising behavior}, and \thrirevision{clear intention during navigation}). \thrirevision{Predictions were made} based on visualizations of observations of the human-robot  interaction, as recorded in our VR navigation dataset. 
Second, we investigate\track{d} how well various supervised learning models do this type of inference in comparison to humans. 
\track{Third}, we stud\track{ied} the generalization capabilities of supervised learning methods to users unseen at training time. 

Our analyses bring understanding to the complexity of predicting humans' \thrirevision{perceptions} of robot performance in navigation tasks  \thri{and enabled us to finally conduct a real-world demonstration in which a robot uses a machine learning model to predict
how a human perceives it in a university campus. 
We conclude this paper by discussing the implications of our results for implementing autonomous systems that infer human perceptions of robot performance using implicit feedback in real-world navigation scenarios. We hope that our recommendations facilitate future efforts to make robots more aware of their failures during navigation \cite{tian2021taxonomy}, as well as facilitate aligning robot behavior to human preferences based on implicit feedback \cite{mcquillin2022learning,cui2021empathic,chetouani2021interactive}.
}

\section{Related Work}
\label{sec:related-work}

This section discusses prior work in relation to our contributions. First, we discuss human \thrirevision{perceptions} of robot performance, especially in regards to robot motion. Then, we distinguish between explicit and implicit human feedback, the latter being the focus of our work. Finally, we briefly review data collection methodologies in HRI.

\subsection{\thrirevision{Perceptions} of Robot Performance}
Understanding human \thrirevision{perceptions} of robot performance is important.
The \thrirevision{perceptions} can be used to evaluate robot policies \cite{tan2019one,lo2019perception,pirk2022protocol} and to create better robot behavior  \cite{thomaz2008teachable,mitsunaga2008adapting,cui2021understanding, bera2019improving}, increasing the likelihood of robot adoption.
In this work, we focus on inferring three robot performance dimensions relevant to navigation \cite{gao2022evaluation}: \textit{ competence}, \textit{surprising behavior}, and \textit{clear intent}. Robot competence is a popular performance metric \thrirevision{in robotics}  \cite{carpinella2017robotic}, especially in robot navigation \cite{mavrogiannis2022social,tsoi2021approach,angelopoulos2022you}. \thrirevision{In our work, competent robot navigation behavior corresponds to effectively guiding a human to a destination.}
Surprising behavior violates expectations, \thrirevision{which is often considered undesired  \cite{asavanant2021personal,francis2023principles} and may require explanations by the robot \cite{brandao2021experts}}. Meanwhile, clear \thrirevision{intentions} means the robot enables an observer to infer the goal of its motion \cite{dragan2013legibility}. Prior work suggests that if humans fail to anticipate the  motion of a robot because it acts surprisingly or its intent is unclear, they will likely have trouble coordinating their own behavior with it~\cite{sciutti2018humanizing, dragan2015effects}. \thrirevision{There are other perceptions about a robot beyond robot competence, surprising behavior, and clear intent that one may want to model in Human-Robot Interaction, like human perceptions of  discomfort with a robot \cite{carpinella2017robotic,kidokoro2013will} or perceived safety \cite{rubagotti2022perceived,akalin2022you}; however, this is out of the scope of the present work.}

\subsection{Implicit Human Feedback}
We distinguish between explicit and implicit human feedback about robot performance. Explicit feedback corresponds to purposeful or deliberate information conveyed by humans to robots, e.g., through preferences \cite{biyik2022aprel,suresh2023robot} or survey instruments \cite{avrunin2014socially,mavrogiannis2022social}. Meanwhile, implicit feedback are cues and signals that people exhibit without intending to communicate some specific information about robot performance, yet they can be used to infer such perceptions. 
Inferring performance from implicit feedback can reduce the chances of excessively querying users for explicit feedback in robot learning scenarios \cite{rivoire2016delicate,gucsi2020ask}, thereby minimizing the risk of feedback fatigue~\cite{lin2020review}.
Learning from implicit feedback is not without challenges, however, as it can be difficult to interpret~\cite{cui2021empathic, stiber2022modeling}. For example, this can happen due to inter-person variability in facial expressions \cite{gunes2008lab}, similar signals being produced for different reasons~\cite{candon2023nonverbal}, or signals changing over time as interactions progress~\cite{candon2024react}.

Our work considers a variety of nonverbal implicit signals, including gaze, body motion, and facial expressions, 
which have long been studied in social signal processing \cite{vinciarelli2009social}. While in some cases these signals are treated as explicit feedback (e.g., to interrupt an agent  \cite{yan2020frownonerror}), we consider them implicit feedback because we do not prime humans to react in specific ways to a robot. As such, our work is closer to  \cite{cui2021empathic,wachowiak2022analysing,mcquillin2022learning,stiber2022effective,candon2023nonverbal}, which used \track{these} signals to identify critical states during robot operation, detect robot errors, and adjust robot behavior.

\begin{figure}[tb!p]
    \centering
    \includegraphics[width=.97\textwidth]{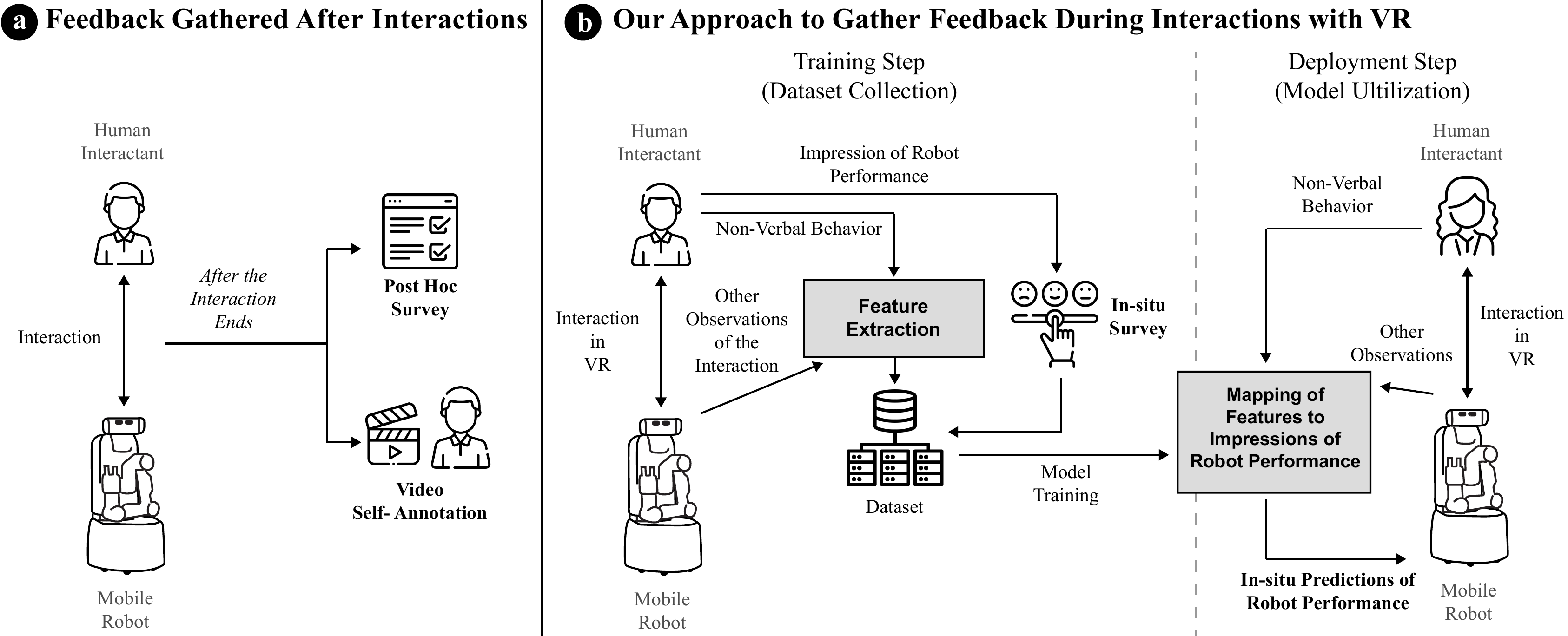}
    \caption{
        a) It is typical to gather explicit human feedback about robot performance using surveys after human-robot interactions conclude because interruptions by the experimenters can easily bias human-robot social encounters. Unfortunately, the feedback from surveys tends to be very limited, making it difficult to understand robot performance at a granular level. Alternatively, participants may complete video annotations of their experiences~\cite{zhang2023sean}, but this can be time consuming and taxing, especially in continuous navigation tasks. 
        b) In this work, we first collect a dataset of human \thrirevision{perceptions} of a robot's performance by prompting participants \textit{during} interactions using VR (Training Step in the diagram). Then, we use this explicit feedback to train  models that infer human \thrirevision{perceptions} of robot performance based on observations of the interactions, especially including observations of human implicit feedback. The value of such a model is that once it is trained, it can be reused to estimate robot performance during new interactions (Deployment Step), without having to ask humans for explicit feedback as in the training step.
    }
    \label{fig:flowchart}
\end{figure}

\subsection{Data Collection in HRI: VR and Other Methodologies}

Different kinds of HRI research methods have been used in the literature to gather interaction data, such as in-person user studies (e.g., \cite{gockley2007natural,trautman2015robot,mavrogiannis2022social}), observational public data collections (e.g., \cite{martin2019jrdb,karnan2022socially}), crowdsourcing studies (e.g., \cite{breazeal2013crowdsourcing,toris2014robot, inamura2021vr}), etc. See \cite{bartneck2020human} for an introduction to these methods.

We considered different ways of conducting our data collection, but ultimately opted for gathering data  with simulated human-robot interactions in VR for several reasons. 
First, in contrast to real-world data collection, simulation facilitated querying humans about their \thrirevision{perceptions} of robot performance during interactions and resulted in fewer negative consequences for interrupting the navigation task. This is illustrated in Fig. \ref{fig:flowchart}. In lab  studies, for instance, surveys that gather general \thrirevision{perceptions} of a robot are typically administered at the end of interactions to avoid interrupting the natural flow of events \cite{zhang2023self}, which can cause unintended effects on collaborative tasks and interactants. In VR simulations, however, we can gather feedback in-situ. We can  freeze time during human-robot interactions, query a participant about their \thrirevision{perceptions} of robot performance through the VR display, and then resume the simulation as if the interruption had not occurred.

Second, \track{we started our research by utilizing VR because}, simulations made interactions safer in contrast to \track{those in the} real-world. The reason is that we wanted to expose participants not only to good robot navigation behavior, but also bad behavior. This was key for inducing a wide range of \thrirevision{perceptions} about robot performance during data collection and, thus, capturing varied implicit feedback. Prior work has used simulations in HRI for safety reasons as well  \cite{mitchell2020safety,huck2021testing}.

Third, in contrast to crowdsourcing data collection procedures, our in-person data collection reduced unrelated participant distractions~\cite{borgo2017crowdsourcing} and minimized potential issues with participant's internet speed \cite{hwang2021ideabot,tsoi2021approach}. Early in our research, we considered using interactive surveys \cite{tsoi2021approach} for our data collection while capturing implicit feedback signals through the webcams of remote participants (e.g., as in \cite{candon2023nonverbal}). However, after testing both this setup and VR, we thought that the increased level of immersion afforded by VR was important to gather naturalistic feedback.

\track{While we opted for using simulations in our work, they are not without limitations. In particular,  simulations can result in a sim-to-real gap, as discussed before in HRI and other robotics areas (e.g., \cite{bharadhwaj2019data,collins2021review,choi2021use,anderson2021sim,li2019comparing,HigginsISER2023}). This gap can emerge in HRI because of differences in physics between simulation and the real-world as well as the human-robot interactions in simulation not reflecting the real-world experience \cite{HigginsISER2023}. Indeed, prior work suggests that virtual robots may be perceived as more discomforting than real robots \cite{li2019comparing}. Thus, towards the end of this paper, we explored applying the insights from our work with VR data to a real-world demonstration, paving the path towards predicting \thrirevision{perceptions} of robot performance in real\thrirevision{-world} application scenarios.}
\thrirevision{Being able to make such predictions 
opens up doors for adapting robot behavior to better align with human's desire (e.g., by treating the predicted human perceptions as a reward signal in reinforcement learning \cite{bai2022training}).}

\section{Problem Statement \& Research Questions}
\label{sec:problem}

We study if a person's \thrirevision{perceptions} of a robot's performance can be predicted using observations of their interaction in dynamic tasks involving navigation. Specifically, we aim to learn a mapping from a sequence of observations to an individual's reported \thrirevision{perceptions} at the end of the sequence (as in Fig.~\ref{fig:flowchart}b). 
We consider multiple robot performance dimensions on a 5-point scale, as detailed later in Sec. \ref{sec:vr}.

Consider a dataset of observations and performance labels, $\mathcal{D} = \{( \mathbf{o}_{1:T}^i,y^i)\}$, where $\mathbf{o}_{1:T}$ is an observation sequence of length $T$, $y$ is a performance rating given by a robot user at the end of the sequence, and $i$ identifies a given data sample.
We place emphasis on predicting a person's \thrirevision{perceptions} of a robot by considering observations of their implicit feedback. Thus, the observations $\mathbf{o}_t^i$  include features that describe the person's non-verbal behavior, such as \track{their motion,} gaze and facial expressions.
Also, the observations include features that describe the spatial behavior of all the agents in the environment, the navigation task, and the space occupied by static objects.
Given this data, we investigate three fundamental research questions:
\vspace{0.5em}
\begin{enumerate}[wide, labelwidth=!, labelindent=0pt,itemsep=0.5em]
\item \textbf{\textit{How well can human observers predict a user's \thrirevision{perceptions} of robot performance?}} By answering this question, we obtain a human baseline for learning a function $f: \mathcal{O}_{1:T} \rightarrow \mathcal{Y}$, where $\mathcal{O}$ is the observation space 
and $\mathcal{Y}$ is performance.  
Also, through this question, we study the impact of two types of observations in the prediction task: observations that describe fine-grained facial expressions for a robot user; and other observations about the user, the robot and their environment.
As mentioned earlier, observations of fine-grained expressions have gained popularity in recent work to infer human perceptions of an agent's behavior \cite{cui2021empathic,candon2023nonverbal,zhang2023self,stiber23hri}.
Other observations (e.g. body motion and nearby static obstacles) can be more easily computed in real-world navigation tasks, but their usefulness on a robot's ability to infer users' \thrirevision{perceptions} of their performance is less understood.

\item \textbf{\textit{Can machine learning methods predict \thrirevision{perceptions} of robot performance as well as humans?}} Ultimately, we are interested in bringing us forward to a future where machine learning models facilitate evaluating robot performance at scale, without having to necessarily ask users all the time for explicit feedback (as in the Deployment Step of Fig.~\ref{fig:flowchart}b). Thus, we evaluate various machine learning models to approximate the function $f$, as defined for the prior question. 

\item \textbf{\textit{How well can machine learning models generalize to unseen users?}} In future robot deployments, a robot may interact with completely new users. Thus, we analyze the performance of various machine learning models in predicting \thrirevision{perceptions} of robot performance according to users for whom the model had no data at training time.
\end{enumerate}

\track{We study the above questions using data from SEAN-VR \cite{zhang2023sean}, as described in the next two sections. Later, in Sec. \ref{sec:rw-demo}, we leverage our findings in VR to create a real-world demonstration through which we investigate  predicting human \thrirevision{perceptions} of robot performance in two university environments.}

\section{Data Collection with SEAN and VR}
\label{sec:vr}

For our \track{VR} data collection, we leveraged the SEAN 2.0 simulator \cite{tsoi2022sean}. SEAN 2.0 integrates with the Robot Operating System (ROS) \cite{quigley2009ros} 
and supports Virtual Reality \cite{zhang2023sean}. Participants used a Vive Pro Eye VR device to control an avatar in a warehouse (as in Fig.~\ref{fig:setup}(a)). The VR headset captured implicit signals from the participants, like eye and lip movements. 

During data collection, the participants had to follow a Fetch robot that guided them to a destination that was unknown to them a priori but was marked by a red cross on the ground. Fig.~\ref{fig:setup}(b) shows a first-person view of the simulation during robot-guided navigation. The Fetch robot was controlled with ROS in SEAN. The  environment contained other algorithmically controlled pedestrians and \track{warehouse} obstacles provided by SEAN 2.0. 

The participants provided ratings of robot performance through the simulation's VR interface. The frame rate of the rendering of the virtual environment in the participants' first-person view in VR was over 30 frames per second. 
Our data collection protocol, described below, was approved by our local Institutional Review Board and refined via pilots. 

\subsection{Participants}
\label{sec:participants}

We recruited 60 participants using flyers and by word of mouth. They were at least 18 years old, fluent in English, and had normal or corrected-to-normal vision. Overall, 19 participants identified as female, 40  as male, and 1  as non-binary or third gender. Most of them were university students, and ages ranged from 18 to 43 years old. Participants were somewhat familiar with robots, as indicated by a mean rating of M = 4.20 (with standard error SE = 0.18) on a 7-point Likert responding format (1 being lowest). Yet, they were somewhat unfamiliar with VR (M = 3.72, SE = 0.20). No participant had prior experience with SEAN or social robot navigation in VR.

\subsection{Data Collection Procedure}
\label{sec:procedure}

\noindent
\textbf{Protocol:}
A data collection session took place as follows. First, the participant provided demographics data. Second, the experimenter introduced the robot, explained the navigation task in which the participant was to follow the robot, and demonstrated how to use the VR device to control their avatar in SEAN and label robot performance. 
Third, the participant experienced four navigation tasks with the robot, each with a particular starting position and destination. For consistency, the pedestrians were controlled using the same behavior graph controller provided in SEAN 2.0 \cite{tsoi2022sean} and the robot used the same navigation logic across the tasks. 

In each task, the robot guided the participant to the destination and repeatedly changed its behavior (as further detailed below). Importantly, the interaction was paused before and after each behavior change took place, at which point the participant was asked to evaluate the robot's most recent navigation performance.
A typical data collection session was completed in 45 min to 1 hour. Participants were compensated US\$15 for their time.

\vspace{0.5em}
\noindent
\textbf{Robot Behaviors:} During a navigation task, the robot switched between one of these three types of behavior:
\begin{description}
[align=left, leftmargin=0em, labelsep=0.3em, font=\normalfont\itshape, itemsep=0.2em, parsep=0em]
    \item [1. Nav-Stack.] The robot navigated efficiently to the destination based on the path planned by the ROS Navigation Stack with social costs \cite{lu2014layered}. The planned paths generally minimized navigation time while avoiding collisions and invading personal space. This behavior lasted 40 seconds.
    \item [2. Spinning.] The robot rotated at its current position, which we expected  to be perceived as if the robot was confused. This behavior lasted 20 seconds. It was implemented by sending angular velocity commands to the robot's motion controller. 
    \item [3. Wrong-Way.] The robot moved in the wrong direction, away from the task's destination, effectively making a mistake during navigation. This behavior lasted 20 seconds and was implemented using the Navigation Stack  with social costs as well, but with an incorrect navigation goal.
\end{description}

Unbeknownst to the participants, the robot switched to \textit{Nav-Stack} behavior after \textit{Spinning} or \textit{Wrong-Way} during navigation. It randomly switched to \textit{Spinning} or \textit{Wrong-Way} after finishing \textit{Nav-Stack}. The design was intended to maintain a consistent rate of sub-optimal behavior and avoid user boredom or significant confusion, which can be caused by more stochastic behavior patterns that are hard for participants to reason about. We expected the behaviors to elicit both positive and negative views of the robot, leading to a large variety of non-verbal reactions and \thrirevision{perceptions} of robot performance.

\vspace{0.5em}
\noindent
\textbf{\thrirevision{Perceptions} of Robot Performance:} 
During a navigation task, we paused the interaction at 4 seconds \textit{before}, and at 8 seconds \textit{after} the robot switched between behaviors. The elapsed time for the latter pause was longer in order to give people enough time to experience the latest robot behavior. 

As shown in the supplementary video, \thrirevision{perceptions} of robot performance were provided through an interface embedded in the simulation. The interface asked the participants to indicate their \thrirevision{perceptions} about the robot’s most recent performance in regard to: 1) \textit{``how competent was the robot at navigating,''} 2) \textit{``how surprising was the robot’s navigation behavior,''} and 3) \textit{``how clear were the robot’s intentions during navigation.''} Participants provided ratings for these three dimensions of robot performance on a 5-point Likert responding format, e.g., with 1 being ``incompetent'', 2 being ``somewhat incompetent'', 3 being ``neither competent nor incompetent'', 4 being ``somewhat competent'', and 5 being ``competent''.

\subsection{Observations} 
 
We organized observations of human-robot interactions, as recorded in SEAN-VR \cite{zhang2023sean}, into the  features described below. More details about these features are provided in the Appendix. 

\begin{description}[style=unboxed,leftmargin=0cm,itemsep=0.5em]
\item [Participants' Facial Expression Features:] We captured the participants' eye and lip movements, as well as their gaze through the VR headset using the VIVE Eye and Facial Tracking (SRanipal) SDK. The eye and lip movements corresponded to 73 features that described the geometry of the face through blend shapes. The gaze was a 3D vector providing the direction of gaze of the person relative to their face.

\item [Spatial Behavior Features:] During navigation, we captured the poses of the robot, the participant, and the other automatically-controlled avatars  on the ground plane of the scene. Then, we computed the  poses of the avatars relative to the robot, considering only those within a 7.2m radius, as this region is typically considered a robot's public space \cite{hall1966hidden,shiomi2010larger,jensen2018knowing}. Each of the features were $(x, y, \theta)$ tuples with $x$, $y$ being the position and $\theta$  the body orientation (yaw angle) relative to a coordinate frame attached to the robot. 

\item [Goal Features:] A navigation task had an associated destination or goal that the robot had to reach. We converted the goal pose in a global frame in the warehouse to a pose in a coordinate frame attached to the robot. This pose described  the robot's proximity and relative orientation to its destination.

\item [Occupancy Features:] During navigation, the robot localized \cite{grisetti2007improved} against a 2-Dimensional (2D) map of the warehouse. We  used a cropped section of the map around the robot (of $7.2$m $\times$ $7.2$m)  to describe the occupancy of nearby  space by static objects.
    
\end{description}

\subsection{Perceived Robot Performance}

\thrirevision{Perceptions} of robot performance were as expected: ratings for competence and clear intention were generally higher for \textit{Nav-Stack} than for \textit{Spinning} and \textit{Wrong-Way}, while the latter two  tended to be more surprising than the former. Pairs of performance dimensions were significantly correlated with absolute Pearson r-values greater than 0.6. An exploratory factor analysis suggested that the dimensions could be combined into one performance factor (which explained 77\% of the variance).

\vspace{0.5em}
Using the features described before and the \thrirevision{perceptions} of robot performance provided by the participants, 
we created a dataset of paired observation sequences and target performance values. We further refer to this data as the SEAN virTual rObot GuidE with impliciT Human fEedback and peRformance Dataset (\textsc{SEAN TOGETHER} Dataset). As  described below, we used this dataset to investigate the research questions in Sec. \ref{sec:problem}.

\section{Findings}

\subsection{How Well Can Human Observers Predict a User's \thrirevision{Perceptions} of Robot Performance?}
\label{sec:human_procedure}

To better understand the complexity of inferring \thrirevision{perceptions} of robot performance, we evaluated how well human annotators could solve the prediction problem. To this end, we administered an online survey through \thrirevision{Prolific,\footnote{\url{www.prolific.co}}} a platform for human data collection and online research studies. In our survey, human annotators observed visualizations of observations in our \textsc{SEAN TOGETHER} Dataset. Then, they tried to predict performance ratings provided by the people who followed the robot.

\vspace{0.5em}
\noindent
\textbf{Method:}
For the survey, we randomly selected 2 data samples from each of the 60 participants in our data collection, with one gathered before and the other gathered after the robot's behavior changed. The observations in each sample corresponded to an 8-second 5-hz window of features right before the corresponding performance label was provided.

\begin{figure}[tb!p]
    \centering
    \includegraphics[width=.5\linewidth]{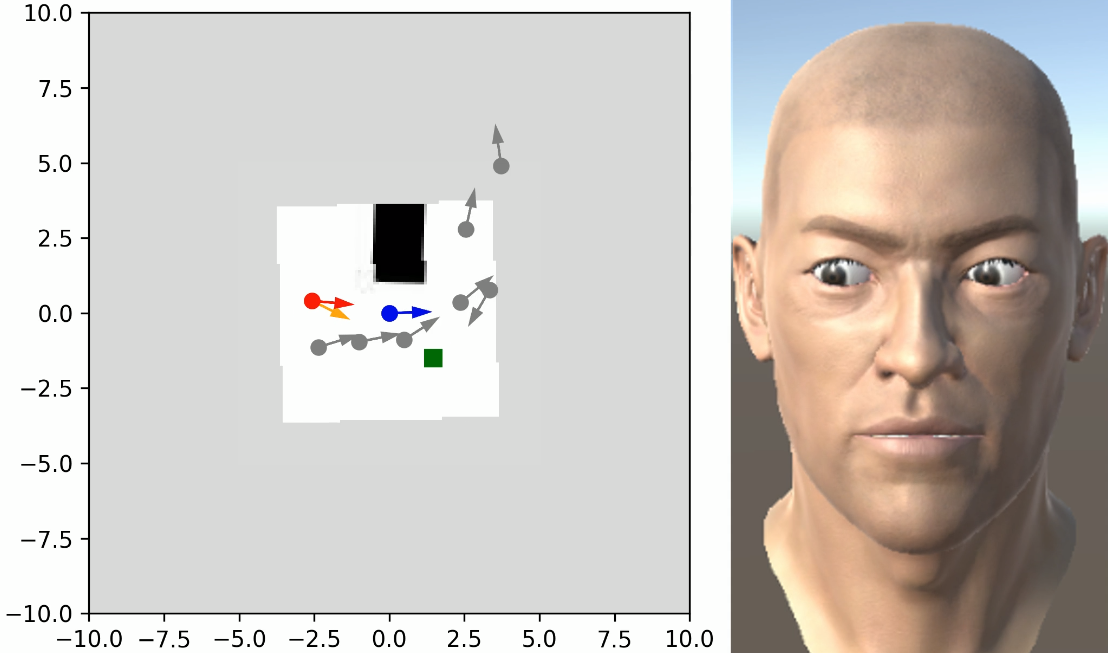}
    \caption{
        A data sample from the \textit{Nav.}+\textit{Facial} condition.
        The \textbf{left} plot shows gaze, spatial behavior, goal, and occupancy features: \includegraphics[]{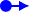} is the robot's pose;
        \includegraphics[]{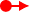} is the pose of the participant following the robot during the VR interaction; 
        \includegraphics[]{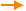} indicates the gaze of the participant;
        \includegraphics[]{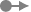} are the poses of algorithmically controlled avatars;
       \includegraphics[]{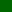} is the destination position that the robot navigated towards;
        and occupancy in the environment is indicated by black pixels (occupied) and white pixels (unoccupied).
        The \textbf{right} visualization shows a rendering of the facial expression features of the participant.
    }
    \label{fig:prolific}
\end{figure}

\vspace{0.5em}
\noindent
As shown in Fig.~\ref{fig:prolific}, data samples were visualized in two ways:
\begin{description}[align=left, leftmargin=0em, labelsep=0.3em, font=\normalfont\itshape, itemsep=0em, parsep=0.2em]
    \item [1. Facial Rendering.]  We created a human face rendering in Unity by replaying the facial expression features on an SRanipal compatible avatar, as shown in Fig.~\ref{fig:prolific} (right). This visualization was motivated by the use of facial expressions in prior work on implicit feedback (e.g.,~\cite{cui2021empathic}).
    \item [2. Navigation Rendering.] We created a plot of features that described the navigation behavior of the robot and the avatars in the simulation. The plot showed features that, using existing perception techniques, may be easier to estimate than facial features in real-world deployments.
    These features are the spatial behavior features, the robot's goal location, the occupied space near the robot, and the gaze direction of the participant -- the last of which could be approximated using an estimate of the person's head orientation \cite{palinko2016robot}. Because prior work suggests that it is easier to make sense of implicit human feedback in context \cite{candon2023nonverbal}, the plot was always centered on the robot, making its surroundings always visible as in Fig.~\ref{fig:prolific} (left).
\end{description}

\begin{figure*}[tb!p]
    \centering
    \includegraphics[fbox=0.5pt 20pt,height=14em]{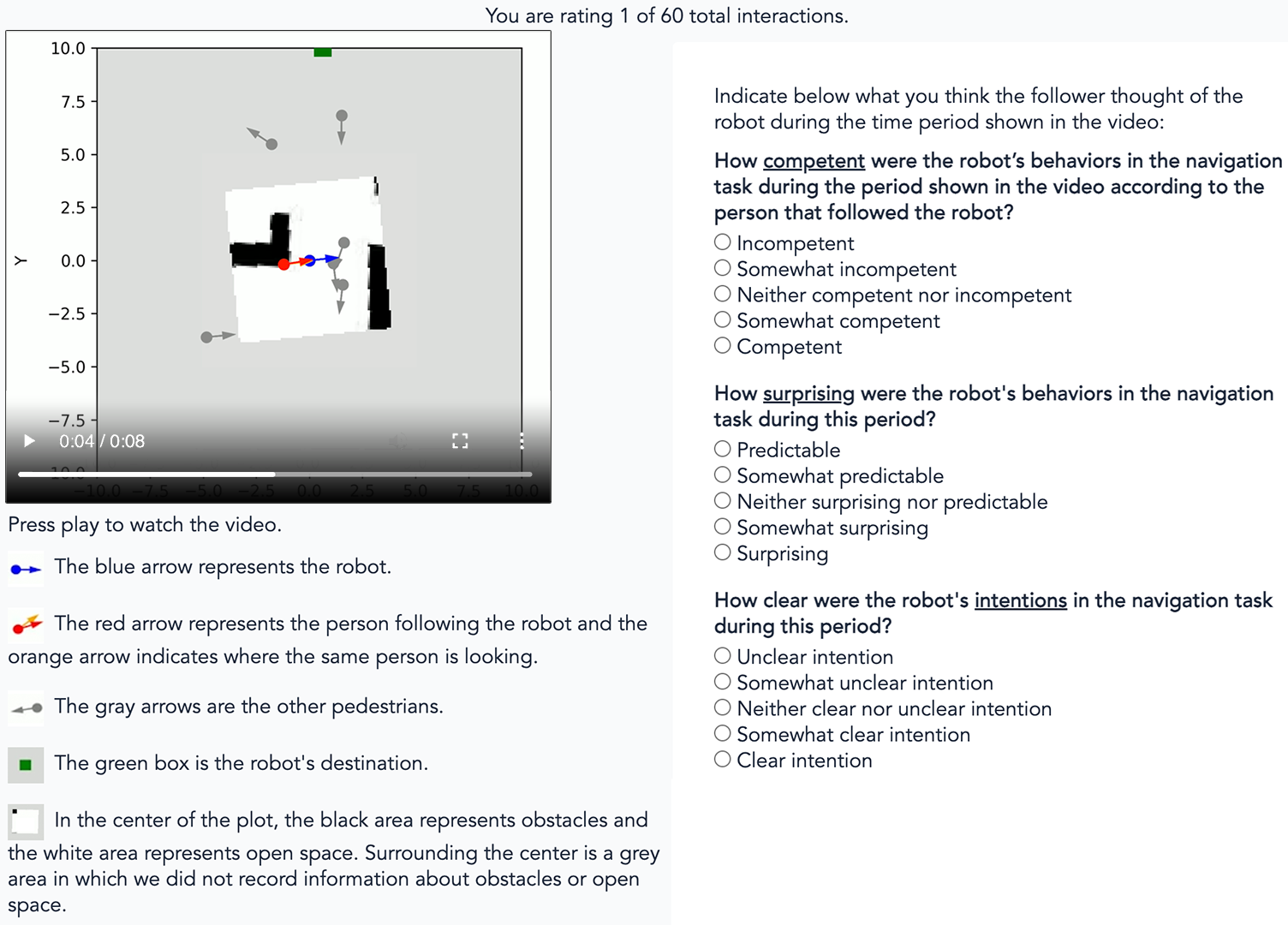}\hfill
    \includegraphics[fbox=0.5pt 20pt,height=14em]{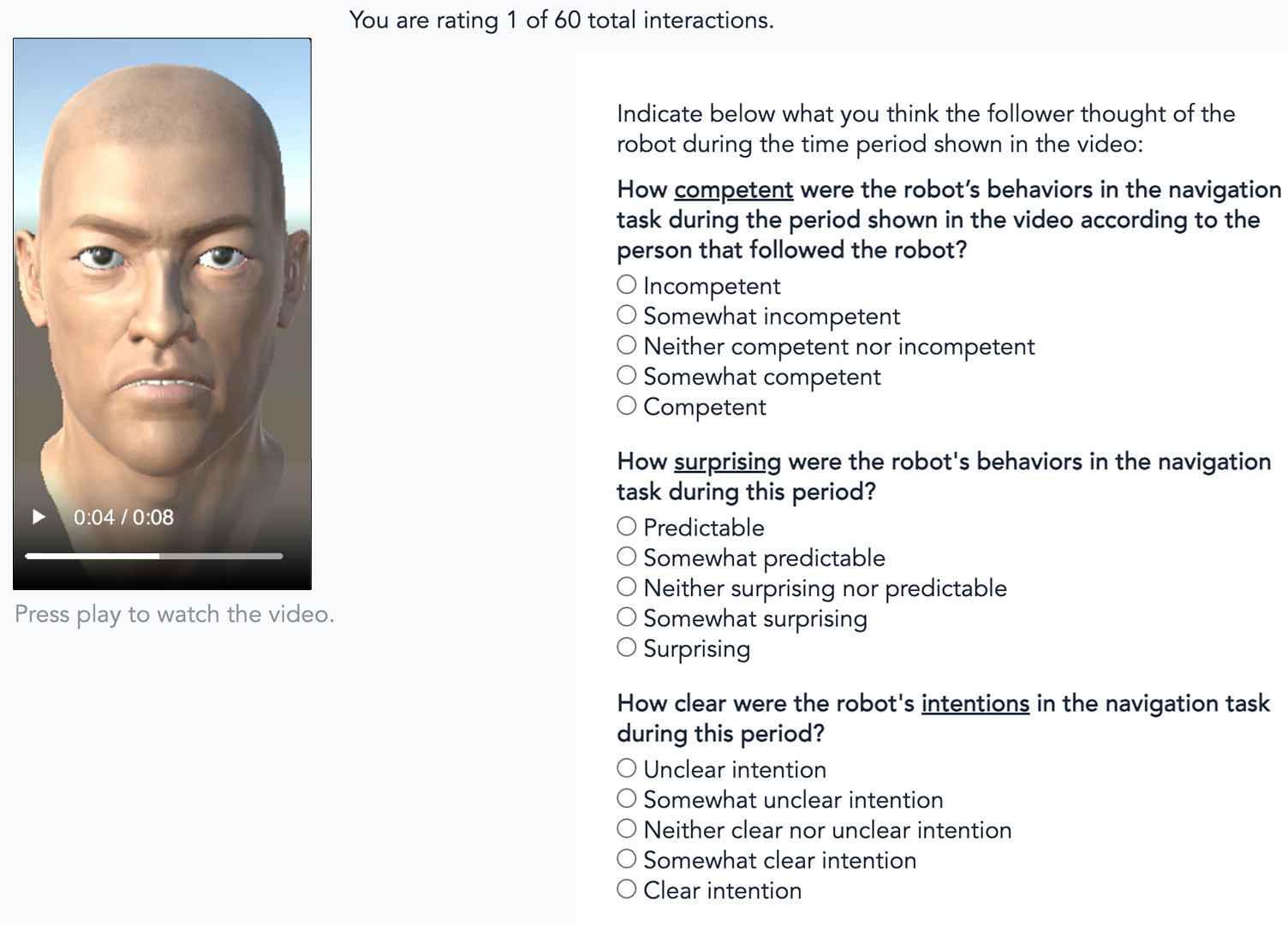}
    \caption{\track{Layout of the interfaces used for video annotation for the human baseline. \textit{Left:} Layout used for the \textit{Nav.-Only} annotation condition, showing the navigation rendering on the left, and questions on the right. \textit{Right:} Layout for the \textit{Facial.-Only}  condition.}}
    \label{fig:spatial_facial_prolific}
\end{figure*}

We used the visualizations to create three annotation conditions that helped understand the value of different features: 
 1) \textbf{\textit{Nav.-Only}}: annotators only saw the navigation rendering \track{(e.g., as in the left image of Fig.~\ref{fig:spatial_facial_prolific})}, \thrirevision{and then completed the annotation};
 2) \textbf{\textit{Facial-Only}}: annotators only saw the facial rendering \track{(e.g., as in the right image of Fig.~\ref{fig:spatial_facial_prolific})}, \thrirevision{and then completed the annotation}; and 3) \textbf{\textit{Nav.+Facial}}: annotators saw the navigation rendering \thrirevision{in the first page}, then the facial rendering \thrirevision{in the second page},  and finally, saw a video with both visualizations \track{next to each other (as in} Fig.~\ref{fig:prolific}) \thrirevision{in the last page and completed the annotation}.

Each of the data samples was annotated by 10 unique people in each condition. The annotators were instructed to predict how the participant who controlled the avatar that followed the robot perceived the robot's performance. The samples they annotated were presented in random order. Each annotator was paid US\$7.5 for approximately $30$ min of annotation time. To encourage 
high-quality annotations, we also gave them a bonus of US\$0.125 for each correct prediction that they made. 

\vspace{0.5em}
\noindent
\textbf{Annotators:} 
We recruited a total of 100 annotators. Thirty-five of them identified as female, 60 as male, and 5 as non-binary or third gender. Ages ranged from 18 to 75 years old. Annotators indicated similar familiarity with robots (M = 4.12, SE = 0.14) as the data collection participants, though the annotators were slightly more familiar with VR  (M = 4.50, SE = 0.16). \track{See the Appendix for details on annotator reliability.}

\vspace{0.5em}
\noindent
\textbf{Results:}
We used linear mixed models estimated with REstricted Maximum Likelihood (REML)~\cite{patterson1975maximum,stroup2012generalized} to analyze errors in the predictions for each performance dimension. 
Our independent variables were  Before/After Robot Behavior Change (\textit{Before}, \textit{After}) and Annotation Condition (\textit{Facial-Only}, \textit{Nav.-Only}, \textit{Nav.+Facial}). Also, we considered Annotator ID as a random effect because annotators provided predictions for multiple data samples. Our dependent variables were the absolute error between an annotator's prediction and the performance rating in our \textsc{SEAN TOGETHER} Dataset. 

\begin{figure}[t!p]
    \centering
    \includegraphics[width=.57\linewidth]{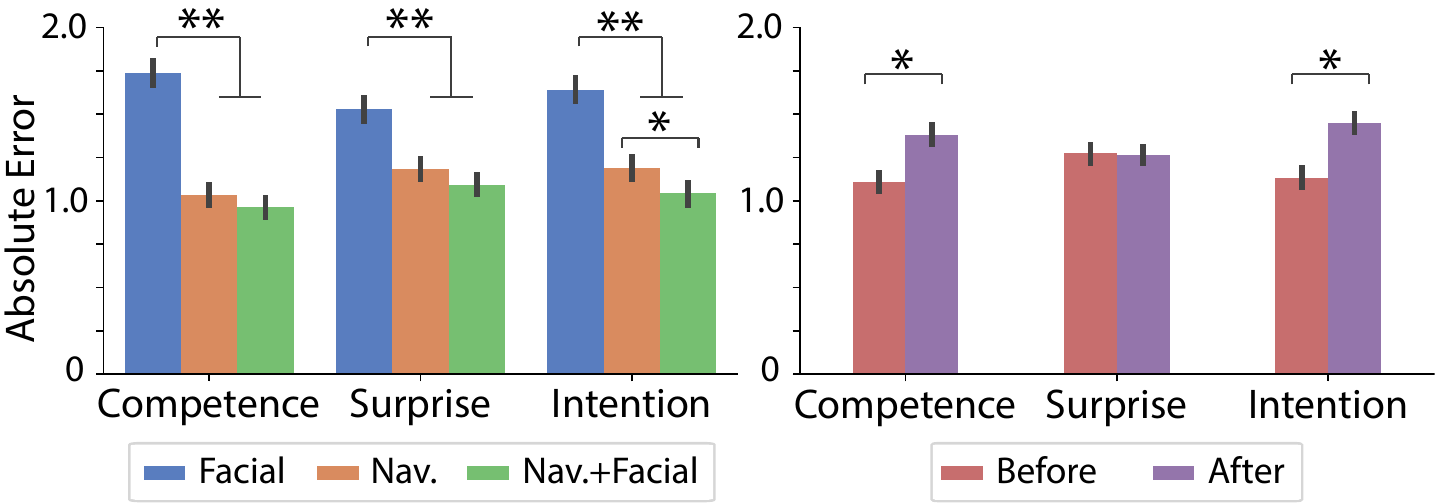}
    \caption{Errors for annotators' predictions by Annotation Conditions (\textit{left}) and Before/After Robot Behavior Change (\textit{right}). (**) and (*) denote $p < 0.0001$ and $p < 0.05$, respectively.}
    \label{fig:human_features_after}
    \vspace{-1.5em}
\end{figure}

We found that the Annotation Condition had a significant effect on the absolute error for Competence, Surprise, and Intention (p $<$ 0.0001 in all cases). As in Fig. \ref{fig:human_features_after} (left), Tukey HSD post-hoc tests showed that for Competence and Surprise, the errors for \textit{Nav.+Facial} and \textit{Nav.-Only} were significantly lower than \textit{Facial-Only}, yet the difference between the former two conditions was not significant.
For Intention, all conditions led to significantly different errors. \textit{Nav.+Facial} resulted in the lowest error, followed by \textit{Nav.-Only} and then  \textit{Facial-Only}.
These results suggest that facial expressions provide information about \thrirevision{perceptions} of robot performance though, more generally, the features used to create the Navigation Renderings seemed to be the most critical for these predictions.

\definecolor{mplBlue}{rgb}{0.7176, 0.8431, 0.9490}
\definecolor{mplOrange}{rgb}{0.9843, 0.8902, 0.8353}

\begin{table}[t!p]
  \caption{Machine learning methods and human annotation (HA) performance on 120 examples. Methods: Random (R) sampling from the distribution of labels in the training set, Random Forest (RF), Multi-Layer Perceptron (MLP), Graph Neural Network (GNN), and Transformer (T). 
  Arrows indicate that higher ($\uparrow$) and lower ($\downarrow$) results are better. Cells with (-) do not have results because a GNN trained on facial features only was effectively an MLP. The \colorbox{mplBlue}{Best} and \colorbox{mplOrange}{Second} results are highlighted.
  }
  \label{tbl:human-results}
  \centering
{

\begin{adjustbox}{width=\textwidth}
\begin{tabular}{rccccccccc}

& \multicolumn{3}{c}{$F_1$-Score ($\mu \pm \sigma$) $\uparrow$}  & \multicolumn{3}{c}{Accuracy ($\mu \pm \sigma$) $\uparrow$}  & \multicolumn{3}{c}{Mean Absolute Error ($\mu \pm \sigma$) $\downarrow$}  \\
\cmidrule(r){2-4} \cmidrule(r){5-7} \cmidrule(r){8-10}
 & Facial & Nav. & Nav.+Facial & Facial & Nav. & Nav.+Facial & Facial & Nav. & Nav.+Facial \\
\cmidrule(r){1-1} \cmidrule(r){2-2} \cmidrule(r){3-3} \cmidrule(r){4-4} \cmidrule(r){5-5} \cmidrule(r){6-6} \cmidrule(r){7-7} \cmidrule(r){8-8} \cmidrule(r){9-9} \cmidrule(r){10-10}
\parbox[t]{7mm}{\multirow{5}{*}{\rotatebox[origin=c]{90}{Competence}}}
HA &   $0.16 \pm 0.0$ &   $0.28 \pm 0.1$ &   $0.29 \pm 0.1$ &  $0.19 \pm 0.1$ &  $0.40 \pm 0.1$ &  $0.42 \pm 0.1$ &   $1.74 \pm 0.2$ &   $1.03 \pm 0.3$ &   $0.99 \pm 0.4$\\ 
R &   $0.18 \pm 0.0$ &   $0.19 \pm 0.0$ &   $0.17 \pm 0.0$ &  $0.21 \pm 0.0$ &  $0.21 \pm 0.0$ &  $0.20 \pm 0.0$ &   $1.73 \pm 0.1$ &   $1.75 \pm 0.1$ &   $1.81 \pm 0.1$\\ 
RF &   $0.19 \pm 0.0$ &   \cellcolor{mplBlue} $0.37 \pm 0.0$ &   \cellcolor{mplBlue} ${0.38 \pm 0.0}$ &  \cellcolor{mplBlue} ${0.33 \pm 0.0}$ &  \cellcolor{mplBlue} ${0.52 \pm 0.0}$ &  \cellcolor{mplBlue} ${0.52 \pm 0.0}$ &   \cellcolor{mplBlue} ${1.43 \pm 0.0}$ &   \cellcolor{mplBlue} ${0.88 \pm 0.0}$ &   \cellcolor{mplBlue} ${0.82 \pm 0.0}$\\ 
MLP &   \cellcolor{mplBlue} ${0.23 \pm 0.0}$ &   $0.29 \pm 0.1$ &   $0.25 \pm 0.1$ &  $0.28 \pm 0.0$ &  \cellcolor{mplOrange} $0.48 \pm 0.0$ &  \cellcolor{mplOrange} $0.44 \pm 0.1$ &   $1.66 \pm 0.1$ &   $1.07 \pm 0.3$ &   $1.19 \pm 0.4$\\ 
GNN &   - &   $0.31 \pm 0.1$ &   $0.33 \pm 0.0$ &  - &  $0.43 \pm 0.1$ &  $0.39 \pm 0.1$ &   - &   $1.22 \pm 0.3$ &   $1.04 \pm 0.0$\\ 
T &   \cellcolor{mplOrange} $0.21 \pm 0.0$ &   \cellcolor{mplOrange} $0.33 \pm 0.0$ &   \cellcolor{mplOrange} $0.33 \pm 0.0$ &  \cellcolor{mplOrange} $0.30 \pm 0.0$ &  $0.43 \pm 0.0$ &  $0.41 \pm 0.1$ &   \cellcolor{mplOrange} $1.58 \pm 0.1$ &   \cellcolor{mplOrange} $0.97 \pm 0.0$ &   \cellcolor{mplOrange} $0.95 \pm 0.0$ \\
\cmidrule(r){1-1} \cmidrule(r){2-2} \cmidrule(r){3-3} \cmidrule(r){4-4} \cmidrule(r){5-5} \cmidrule(r){6-6} \cmidrule(r){7-7} \cmidrule(r){8-8} \cmidrule(r){9-9} \cmidrule(r){10-10}
\parbox[t]{7mm}{\multirow{5}{*}{\rotatebox[origin=c]{90}{Surprise}}} 
HA &  $0.18 \pm 0.0$ &  $0.24 \pm 0.1$ &  $0.25 \pm 0.1$ &   $0.20 \pm 0.1$ &   $0.30 \pm 0.1$ &   $0.32 \pm 0.1$ &  $1.53 \pm 0.3$ &  $1.19 \pm 0.2$ &  $1.12 \pm 0.2$\\ 
R &  $0.19 \pm 0.0$ &  $0.21 \pm 0.0$ &  $0.17 \pm 0.0$ &  $0.20 \pm 0.0$ &   $0.21 \pm 0.0$ &   $0.18 \pm 0.0$ &  $1.64 \pm 0.1$ &  $1.60 \pm 0.1$ &  $1.68 \pm 0.1$\\
RF &  \cellcolor{mplBlue}  ${0.29 \pm 0.0}$ &  \cellcolor{mplBlue}  ${0.38 \pm 0.0}$ &  \cellcolor{mplBlue} ${0.34 \pm 0.0}$ &   \cellcolor{mplBlue} ${0.30 \pm 0.0}$ &   \cellcolor{mplBlue} ${0.40 \pm 0.0}$ &   \cellcolor{mplBlue} ${0.34 \pm 0.0}$ &  \cellcolor{mplOrange} $1.30 \pm 0.0$ &  \cellcolor{mplBlue} ${0.93 \pm 0.0}$ &  \cellcolor{mplBlue} ${0.98 \pm 0.0}$\\ 
MLP &  $0.24 \pm 0.0$ &  $0.26 \pm 0.1$ &  $0.24 \pm 0.1$ &   $0.25 \pm 0.0$ &   $0.30 \pm 0.0$ &   $0.29 \pm 0.1$ &  \cellcolor{mplBlue} ${1.23 \pm 0.1}$ &  $1.12 \pm 0.2$ &  $1.08 \pm 0.1$\\ 
GNN &  - &  $0.29 \pm 0.0$ &  $0.27 \pm 0.0$ &   - &   $0.30 \pm 0.0$ &   $0.28 \pm 0.0$ &  - &  $1.13 \pm 0.1$ &  $1.07 \pm 0.1$\\ 
T &  \cellcolor{mplOrange} $0.27 \pm 0.0$ &  \cellcolor{mplOrange} $0.29 \pm 0.0$ &  \cellcolor{mplOrange} $0.32 \pm 0.1$ &   \cellcolor{mplOrange} $0.28 \pm 0.0$ &   \cellcolor{mplOrange} $0.31 \pm 0.0$ &   \cellcolor{mplOrange} $0.33 \pm 0.1$ &  $1.37 \pm 0.1$ &  \cellcolor{mplOrange} $1.07 \pm 0.1$ &  \cellcolor{mplOrange} $1.04 \pm 0.1$ \\
\cmidrule(r){1-1} \cmidrule(r){2-2} \cmidrule(r){3-3} \cmidrule(r){4-4} \cmidrule(r){5-5} \cmidrule(r){6-6} \cmidrule(r){7-7} \cmidrule(r){8-8} \cmidrule(r){9-9} \cmidrule(r){10-10}
\parbox[t]{7mm}{\multirow{5}{*}{\rotatebox[origin=c]{90}{Intention}}}
HA &   $0.18 \pm 0.0$ &   $0.25 \pm 0.1$ &  \cellcolor{mplOrange} $0.30 \pm 0.1$ &  $0.21 \pm 0.1$ &  $0.37 \pm 0.2$ &  \cellcolor{mplBlue}  ${0.41 \pm 0.1}$ &   $1.64 \pm 0.2$ &   \cellcolor{mplOrange} $1.19 \pm 0.4$ &   \cellcolor{mplBlue} ${1.07 \pm 0.2}$\\ 
R &   $0.21 \pm 0.1$ &   $0.19 \pm 0.0$ &   $0.17 \pm 0.0$ &  $0.23 \pm 0.1$ &  $0.22 \pm 0.0$ &  $0.19 \pm 0.0$ &   $1.70 \pm 0.1$ &   $1.73 \pm 0.1$ &   $1.80 \pm 0.1$\\ 
RF &   \cellcolor{mplBlue} ${0.28 \pm 0.0}$ &   \cellcolor{mplOrange} $0.28 \pm 0.0$ &   $0.24 \pm 0.0$ &  \cellcolor{mplBlue} ${0.37 \pm 0.0}$ &  \cellcolor{mplBlue} ${0.43 \pm 0.0}$ &  \cellcolor{mplOrange} $0.41 \pm 0.0$ &   \cellcolor{mplBlue} ${1.45 \pm 0.0}$ &   \cellcolor{mplBlue} ${1.13 \pm 0.0}$ &   \cellcolor{mplOrange} $1.14 \pm 0.0$\\ 
MLP &   \cellcolor{mplOrange} $0.27 \pm 0.0$ &   $0.26 \pm 0.1$ &   $0.22 \pm 0.0$ &  $0.31 \pm 0.0$ &  $0.41 \pm 0.1$ &  $0.39 \pm 0.1$ &   $1.86 \pm 0.1$ &   $1.31 \pm 0.3$ &   $1.51 \pm 0.5$\\ 
GNN &   - &   $0.28 \pm 0.0$ &   $0.29 \pm 0.0$ &  - &  $0.37 \pm 0.0$ &  $0.35 \pm 0.0$ &   - &   $1.32 \pm 0.1$ &   $1.25 \pm 0.1$\\ 
T &   $0.24 \pm 0.0$ &   \cellcolor{mplBlue} ${0.29 \pm 0.1}$ &   \cellcolor{mplBlue} ${0.32 \pm 0.0}$ &  \cellcolor{mplOrange} $0.33 \pm 0.0$ &  \cellcolor{mplOrange} $0.41 \pm 0.0$ &  $0.40 \pm 0.0$ &   \cellcolor{mplOrange} $1.63 \pm 0.1$ &   $1.21 \pm 0.1$ &   $1.20 \pm 0.1$ \\

\end{tabular}
\end{adjustbox}
}
\vspace{1.5em}
\end{table}

\begin{figure}[b!p]
    \centering
    \includegraphics[width=.7\textwidth]{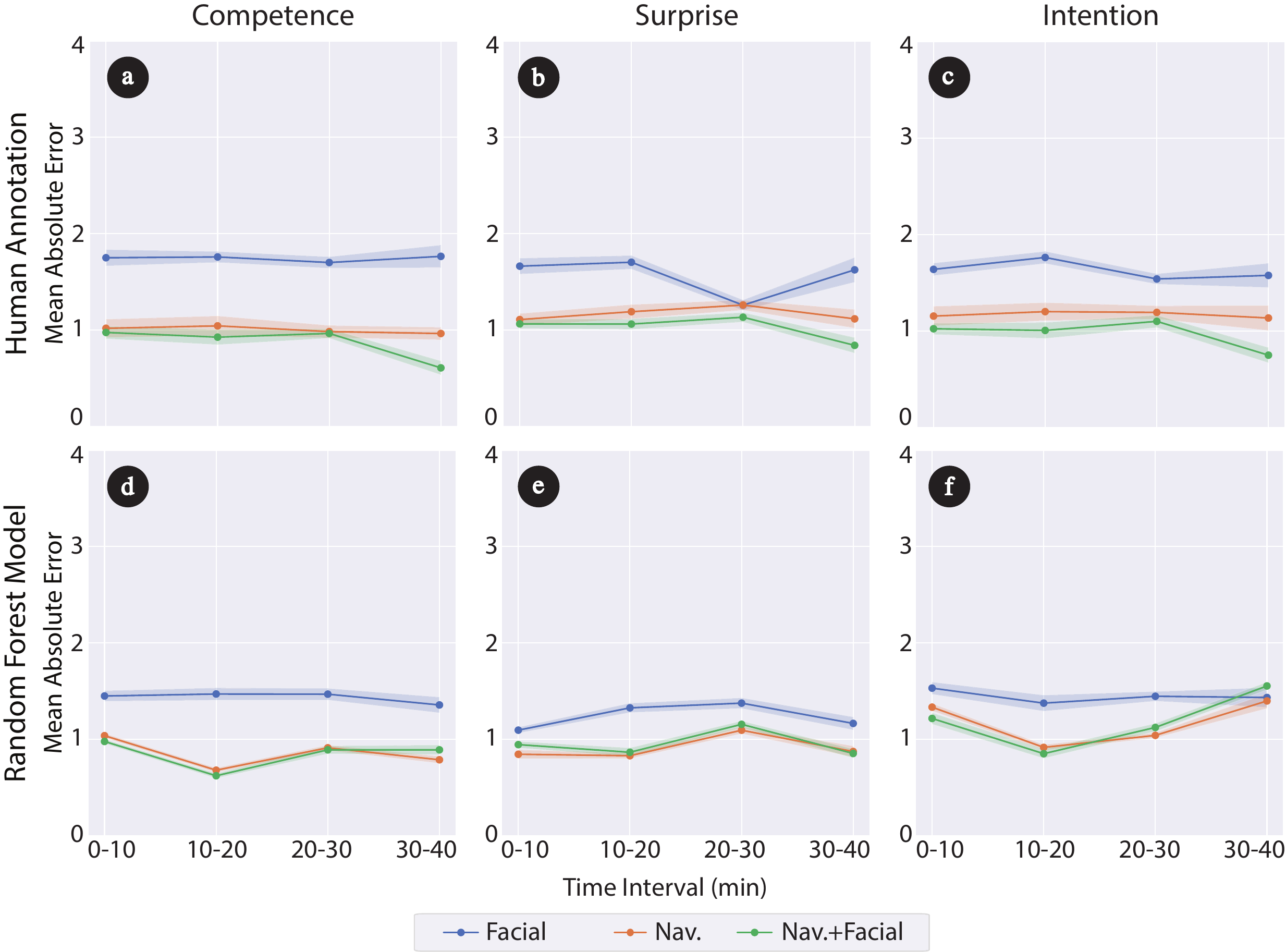}
    \caption{
        Mean Absolute Errors (MAE) of human annotation and Random Forest (RF) results over 10-minute intervals of the data collection sessions. MAE was computed for all data samples in each interval, and then the average and standard errors of MAE were calculated considering the performance of the 10 unique annotators (for human annotation results in (a)--(c)) or the 10 Random Forest models trained with different seeds in Table~\ref{tbl:human-results} (RF results in (d)--(f)).
    }
    \label{fig:time_vs_l1}
\end{figure}

Before/After Robot Behavior Change had a significant effect on the prediction errors for Competence and Intention (p $<$ 0.0001 in both cases). As in Fig. \ref{fig:human_features_after} (right), the error was significantly lower for samples \textit{Before} a behavior change than for samples \textit{After} a change for these performance dimensions. We suspect this was because the robot sometimes demonstrated two behaviors in the samples collected \textit{After} a behavior change, but in the case of \textit{Before} behavior change, the robot only showed one behavior making these data samples more consistent and easier to reason about.

Table \ref{tbl:human-results} shows the $F_1$-Scores for the annotator predictions  (see HA rows). The low $F_1$ scores suggest that correctly predicting \thrirevision{perceptions} of robot performance on a 5-point responding format was difficult for humans. Despite this, we suspected that humans could do a more reasonable job at distinguishing \thrirevision{perceptions} of poor robot performance from other \thrirevision{perceptions}. If this was the case, then this could open up doors in the future to using this binary signal (instead of the more fine-grained feedback) as a reward signal to adapt robot behavior in navigation tasks, e.g., in line with \cite{knox2009interactively,macglashan2017interactive}. Thus, we transformed the ground truth ratings from our data collection to binary values, one corresponding to low performance (e.g., 1-2 ratings for competence) and another to medium-to-high performance (3-5 ratings for competence). Also, we transformed the annotators' predictions similarly. 
This led to $F_1$ scores of 0.69 for Competence, 0.64 for Surprise, and 0.69 for Intention. As expected, human annotators were better at telling the directionality of robot performance ratings than at predicting their exact magnitude.

Finally, we investigated the performance of human annotations over the span of data collection because prior work suggests that the expressiveness of people engaged in human-robot interactions can change over time \cite{candon2024react}, e.g., potentially due to changes in their expectations about the robot or due to fatigue. 
Figures~\ref{fig:time_vs_l1}(a)--(c) show the evolution of mean absolute errors for the human annotators' predictions over 10-minute intervals of interaction, considering each performance dimension. In general, human performance was very stable, suggesting no major bias over time in participant's spatial behavior or facial expressions. Interestingly, the results also suggested that improvements in performance with an individual feature did not necessarily translate in improvements on the \textit{\textit{Nav.}+\textit{Facial}} condition. Humans may have combined the information from the different implicit feedback modalities in subtle ways when making their predictions about how participants in VR perceived the robot.

\subsection{Can Machine Learning Methods Predict \thrirevision{Perceptions} of Robot Performance as Well as Humans?}
\label{sec:analysis-ml-vs-human}

We compared human prediction performance with a variety of classifiers, including a random forest and neural networks. 

\vspace{0.5em}
\noindent
\textbf{Method:}
Machine learning (ML) models were evaluated on the same samples shown to the human annotators ($n=120$). The rest of the data was used for training ($n=2280$) and validation ($n=569$). 
We trained one model for each combination of feature sets shown to the human annotators (\textit{Facial-Only}, \textit{Nav.-Only}, and \textit{Nav.+Facial}).
The \textit{Nav.} feature set included occupied space near the robot, which we encoded using a ResNet-18 representation \cite{he2016deep}.
We repeated training for each model 10 times with varying random seeds.
The Random Forest (RF) used 100 trees and the depth was grown until leaves had less than 2 samples.
The neural networks had a number of parameters on the same order of magnitude: $\num{5.4e6}$ for a Multi-Layer Perceptron (MLP), $\num{2.1e6}$ for a message-passing Graph Neural Network (GNN) \cite{battaglia2018relational}, and $\num{6.5e6}$ for a Transformer (T) \cite{vaswani2017attention}. Networks were trained using 
minibatch gradient descent with the Adam optimizer and cross-entropy loss. Learning rate, batch size, and dropout 
were chosen using grid search with validation-based early stopping~\cite{prechelt2002early}. 
We also compared all these models with a random sampling baseline.

\vspace{0.5em}
\noindent
\textbf{Results:}
As is shown in Table \ref{tbl:human-results}, ML models outperformed both human-level performance and random baseline in all cases when measured via $F_1$-Score. When measured using Accuracy and Mean Absolute Error, ML models performed the best, except for Intention when using \textit{Nav.+Facial} features. These outcomes indicate that our implicit feedback data contained useful information that can be leveraged by ML models to predict users' \thrirevision{perceptions} of robot performance. Further, ML models trained with \textit{Nav.-Only} and \textit{Nav.+Facial} features 
outperformed those trained with \textit{Facial-Only} features. This finding aligns with our observation in Sec. \ref{sec:human_procedure} on the criticality of the \textit{Nav.} features in comparison to 
the \textit{Facial} features on performance prediction. 

Figures~\ref{fig:time_vs_l1}(d)--(f) show the evolution of mean absolute errors for the Random Forest model, which generally performed the best, over 10-minute intervals of interaction during the data collection. Similar to the results from human annotators (Figures~\ref{fig:time_vs_l1}(a)--(c), Sec.~\ref{sec:human_procedure}), the error for the RF model did not fluctuate drastically, although the performance for Intention prediction with \textit{Nav.} and \textit{Nav.+Facial} features decreased in the last two time intervals of data collection (having higher mean absolute error). The decrease in performance could be the result of a distribution shift, especially in the last interval which had the fewest number of samples because not all interactions took the full 40 minutes. Also, a good proportion of the samples in the last time interval showed the end of navigation tasks, 
at which point the participants could have been more sensitive to robot navigation in the wrong direction. Indeed, there was a higher proportion of lower ratings for Intention in the last interval than in the other intervals, as shown in the Appendix.

\track{To better understand differences in the prediction performance between ML and human annotators, we first identified the examples annotated by humans for which there was a difference greater than 1 in Mean Absolute Error between human annotators and the RF model that tended to perform best. Then, we inspected the 8-second navigation renderings of these data examples, as in Fig.~\ref{fig:spatial_facial_prolific} (left). Among examples where the RF model performed better than humans, 64\% exhibited a major behavior pattern for the robot that persisted despite minor deviations. For example, the robot navigated effectively to the goal most of the time, but was occasionally blocked and had to move around the obstacles. We hypothesize that ML did better in these cases because machine learning can leverage regularities in the data when making predictions without potentially getting distracted with the minor deviations. Among the examples where human annotators performed better, 68\% showed the robot exhibiting more than one behavior (\textit{Nav-Stack}, \textit{Spinning}, or \textit{Wrong-Way}) or the interaction involved unconventional reactions from humans, such as people interfering in the navigation task. We suspect that humans were better in these  cases because they can leverage their prior knowledge about the world to better reason about uncommon variations in the data. For the RF, uncommon observations can be out-of-distribution samples that result in more prediction errors, especially considering the limited size of our dataset.}

\track{Taken together, these results} 
motivated us to focus the analysis in the next section on the aggregate, overall results rather than the interval-based results.

\subsection{Can Machine Learning Generalize to Unseen Users?}
\label{sec:analysis-ml-generalization}

\begin{figure}[b!p]
    \centering
    \includegraphics[width=.6\linewidth]{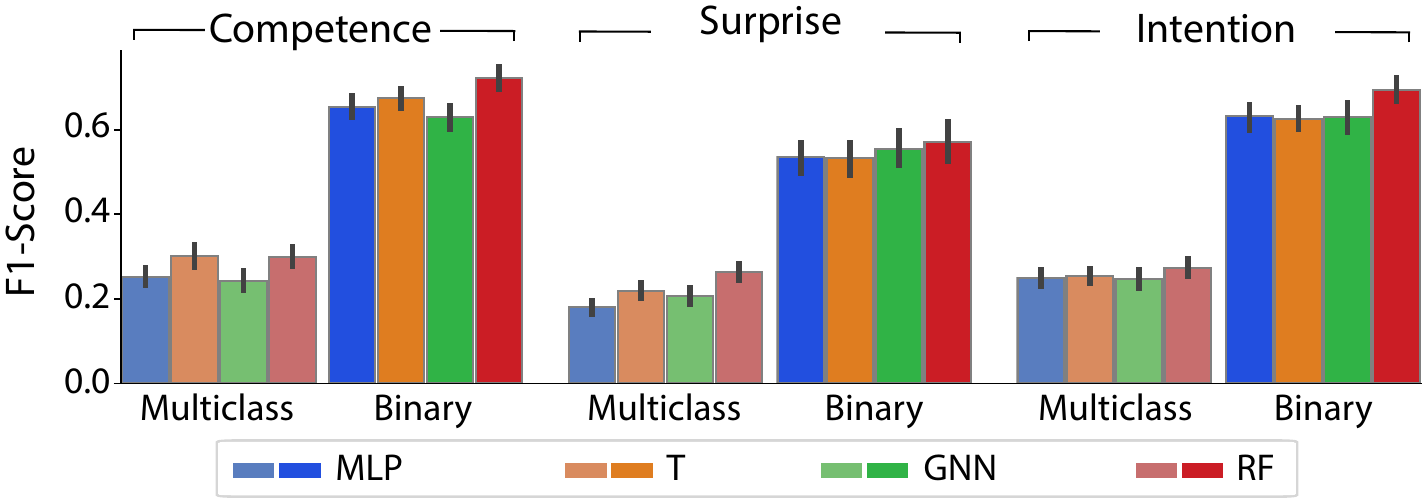}
    \caption{ML models trained on \textit{Nav.$+$Facial} features using leave-one-out cross-validation and evaluated on the held-out participant's data. $F_1$-Scores are computed over 5 classes (Multiclass) and 2 classes (Binary). Error bars represent the standard errors calculated from the $F_1$-Scores per leave-one-out fold. See the text for details.}
    \label{fig:generalization}
\end{figure}

We investigated how well learning models could predict performance by a user whose data was held out from training.

\vspace{0.5em}
\noindent
\textbf{Method:}
We used the models and training scheme from Sec. \ref{sec:analysis-ml-vs-human} with all features (\textit{Nav.+Facial}), but split the data using leave-one-out cross-validation. For each fold, the data for one participant was used as the test set and the remaining examples were split between training ($80\%$) and validation ($20\%$). We searched for new hyperparameters and computed results both on 5-classes and on binary classification. Binary targets and prediction labels were computed as in Sec. \ref{sec:human_procedure}. 

\vspace{0.5em}
\noindent
\textbf{Results:}
Fig. \ref{fig:generalization} reports $F_1$-Scores over all folds. 
The models generalized to unseen people with only a slight reduction in performance in comparison to Table \ref{tbl:human-results}. 
Also, the average $F_1$-Score across all performance dimensions improves from 0.25 in the multiclass case to 0.62 in the  binary case. This makes the ML predictions more usable in practice. 
For example, in the future, we envision deploying the trained ML on new users (as in Fig.~\ref{fig:flowchart}b) in order to detect low robot performance. This could be an indication that the robot made a mistake, triggering interaction recovery behaviors like apologies or explanations \cite{tian2021taxonomy}, which could increase trust on the system \cite{che2020efficient}.

\section{\track{Real-World Demonstration}}
\label{sec:rw-demo}

\track{To investigate whether we could predict human \thrirevision{perceptions} of robot performance in other, more realistic scenarios than those observed in our VR data collection, we conducted a real-world demonstration with a modified Pioneer 3-DX mobile base. More specifically, we conducted a data collection with the mobile robot in two semi-public indoor environments of Yale University, and analyzed how well a random forest model could predict human \thrirevision{perceptions} of robot performance in the real-world setup. This real-world data collection, as further described below, was approved by our local Institutional Review Board.}

\track{The system that we built for real-world data collection was designed in consideration of: 1) we wanted to induce naturalistic interactions between the robot and pedestrians; and 2) we wanted to support the same data collection protocol used with SEAN, as in Section \ref{sec:procedure}.
Therefore, we did not recruit participants prior to the data collection. 
Instead, we operated the robot and, as pedestrians walked nearby, we asked them if they would be willing to follow the robot for a short period and answer brief surveys. In total, \thri{45} pedestrians agreed to follow the robot for this demonstration.}

\begin{figure}[bt!p]
    \centering
    \includegraphics[width=.7\linewidth]{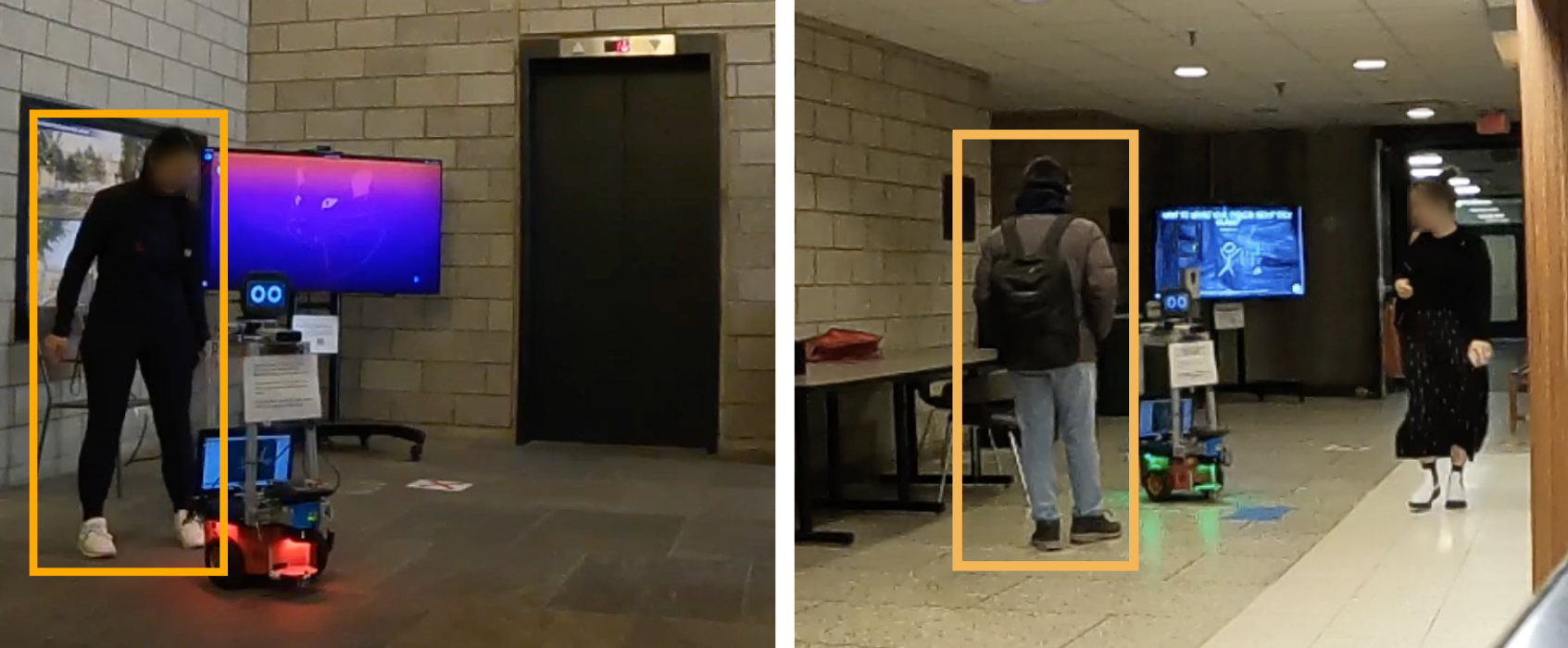}
    \caption{\track{Real-world data collection in two indoor spaces of Yale University. The orange box highlights the follower, i.e., the person that followed the robot during navigation tasks. Other people could pass by the follower and the robot as in the \textit{right} image during data collection. The robot had lights to indicate when it was navigating (green, \textit{right} image) or had paused navigation (red, \textit{left} image).
}}
    \label{fig:real}
\end{figure}

\vspace{0.5em}
\noindent
\textbf{\track{Mobile Robot:}}
\track{The Pioneer 3-DX robot is a differential-drive mobile base and, thus, it moves in a similar way to the Fetch robot used in our VR data collection. We added to the Pioneer robot lights that illuminated green to indicate that it was navigating towards a location, and red to indicate that it had paused navigation. Over the Pioneer base, we built a frame that held a robotic screen face (similar to \cite{vazquez2020gaze,lew2023shutter}) on the very top of the robot, which allowed to easily distinguish the front of the platform. The frame also held two  Kinect Azure RGB-D cameras right below the robot head. Each camera had a 120-degree field of view. One was pointed forward and the other was pointed backwards, which allowed the robot to track people in front and behind it using the Kinect SDK. Additionally, the bottom section of the frame held a 2D LMS-100 Sick LiDAR and a gaming laptop with an Intel Core i7-8750H CPU, 32 GiB of RAM, and an Nvidia GeForce GTX 1070 GPU. The laptop ran the Robot Operating System 
to control the robot using the ROS navigation stack~\cite{quigley2009ros} with social cost layers~\cite{lu2014layered}, which enabled the robot to avoid collisions with nearby people.
Fig.~\ref{fig:real} and our supplementary video show the robot in this demonstration effort. 
}

\vspace{0.5em}
\noindent
\textbf{\track{Demonstration Protocol:}}
\track{We waited for pedestrians to walk by the robot in two locations on a university campus. One location was a subterranean pedestrian tunnel or concourse; the other one was an L-shaped entrance corridor to a building. When pedestrians passed by, we asked them if they would be interested in following the robot as it navigated to a nearby goal marked by a red cross on the ground. For those that agreed, we instructed them that the robot would navigate when it showed a green light. After short intervals of time, it would pause navigation, showing a red light, and they would be asked a few quick questions about their \thrirevision{perceptions} of the action that the robot just performed using a mobile device. The device showed the same questions about robot competence, surprising behavior and clear intent (on a 5-point Likert responding format) as in our VR data collection. Also, the robot navigation behaviors and the timing of questions about robot performance matched those in Sec.~\ref{sec:procedure}.}

\vspace{0.5em}
\noindent
\textbf{\track{Data:}}
\track{
We focused on capturing \textit{Nav.-Only} features (that described the navigation behavior of the robot and humans, as in Sec.~\ref{sec:human_procedure}) for two reasons. First, our prior results with VR data suggested facial expression features were not as critical to make predictions over human \thrirevision{perceptions} of robot performance than the other features. Second,  facial expressions were often occluded, providing no information to the robot.} \thri{In total, we collected 235 examples from this real world demonstration, each consisting of \textit{Nav.-Only} features and associated survey responses.\footnote{\thri{The real-world data that we collected from this demonstration is available at: \url{https://sean-together.interactive-machines.com/}.}}
}

\vspace{0.5em}
\noindent
\textbf{\track{ML Models:}}
\track{Our primary aim was to 
understand the applicability of our approach to infer \thrirevision{perceptions} of robot performance in the real world. However, there were important differences in our VR and real-world data collection setups as a result of real-world constraints. For example,
the real robot had a more limited field of view compared to the simulation where the ground truth motion for all people in the environment was available.
Moreover, the real-world environments were less densely populated than simulation.}

\track{Therefore, to fairly compare our results across simulation and the real world, we trained two types of Random Forest classifiers, given that the RF model generally performed best in Table \ref{tbl:human-results}. One type of model was trained using VR data but we limited the field of view of the robot to 120-degrees forward and backward as well as the maximum number of nearby people input to the model to five individuals. The other type of Random Forest model (with the same parameters) was trained using real-world data. Both types of models were trained considering 5-classes, with binary targets and prediction tables being computed as in Sec. \ref{sec:human_procedure}. 
}

\begin{table}[t!p]
\centering
\caption{
\track{
$F_1$-Score ($\mu \pm \sigma$) for Random Forest models trained using \textit{Nav.-Only} features 
from either the \textit{Real}-world data, or \textit{VR} data considering the nearest 5 people to the robot (as explained in Sec.~\ref{sec:rw-demo}) 
Results include multi-class classification based on the 5-point Likert responses (\textit{Multi-cls}) and binary  classification (\textit{Binary}). Column 1 corresponds to  training on VR data and evaluating on VR data (VR$\rightarrow$VR), Column 2 corresponds to training on VR data and evaluating on real data (VR$\rightarrow$Real), and  Column 3 is training and evaluating on real data (Real$\rightarrow$Real).
}}
\label{tbl:real}
\begin{tabular}{rrccc}
& & \track{(1) VR$\rightarrow$VR} & \track{(2) VR$\rightarrow$Real} & \track{(3) Real$\rightarrow$Real} \\ 
\cmidrule(r){2-2} \cmidrule(r){3-5} 
\parbox[t]{1mm}{\multirow{3}{*}{\rotatebox[origin=c]{90}{\track{Multi-cls}}}} & \track{Competence} & \track{$ 0.30 \pm 0.09$} & \thri{$ 0.21 \pm 0.18$} & \thri{$ 0.27 \pm 0.35$} \\ 
& \track{Surprise} & \track{$ 0.27\pm 0.08$} & \thri{$ 0.26 \pm 0.21$} & \thri{$ 0.26 \pm 0.27$} \\ 
& \track{Intention} & \track{$0.26 \pm 0.08$} & \thri{$ 0.20 \pm 0.28$} & \thri{$ 0.24 \pm 0.34$} \\ 
\cmidrule(r){2-2} \cmidrule(r){3-5}
\parbox[t]{1mm}{\multirow{3}{*}{\rotatebox[origin=c]{90}{\track{Binary}}}} & \track{Competence} & \track{$0.69 \pm 0.10$} & \thri{$ 0.56 \pm 0.41 $} & \thri{$ 0.61 \pm 0.34$} \\ 
& \track{Surprise} & \track{$0.59 \pm 0.18$} & \thri{$ 0.58 \pm 0.36 $} & \thri{$ 0.58 \pm 0.33$} \\ 
& \track{Intention} & \track{$0.65 \pm 0.08$} & \thri{$ 0.55 \pm 0.40$} & \thri{$ 0.60 \pm 0.40$} \\ 

\end{tabular}
\end{table}

\vspace{0.5em}
\noindent
\textbf{\track{Results:}}
\track{
Table \ref{tbl:real} shows the $F_1$-Score  of models evaluated  on the same type of data they were trained on (Sim or Real). For these results, we used leave-one-person-out cross-validation to train and evaluate generalization to new robot followers. That is, data from one person was held out for each fold. Also, Table \ref{tbl:real} shows the performance of the model trained in simulation on real-world data. In this case, a RF model was trained using all the VR data from the VR$\rightarrow$VR case, and then evaluated on the test set for the leave-one-person-out folds for the real-world data. As one would naturally expect based on our prior results with VR data, binary classification resulted in higher performance than multi-class classification in all these cases.}

\track{
In general, performance was higher for models trained and evaluated in simulation (Column 1), which could be the result of having more VR data than real-world data. The results for models trained and evaluated on real data  (Column 3) were close to those that considered  simulation data only (Column 1). This suggested that our methodology to collect real-world data and the RF model are promising for inferring \thrirevision{perceptions} of robot performance in the real world.
}
\thri{Finally, reasonable performance was obtained for the model that was trained with VR data and tested on real-world data (Column 2). This highlights the potential of sim-to-real transfer of machine learning models trained on spatial features as well as the potential of using our VR data to build computational models that predict human perceptions of robot performance in real-world interactions.}

\track{\section{Implications for Real-World Applications}}
We hope that future work leverages our findings to build effective models for mapping implicit human feedback to users' \thrirevision{perceptions} of robot performance in real-world social navigation tasks. To this end, we first recommend prioritizing robust people tracking and pose estimation over computing fine-grained facial expressions, especially when computational resources may be limited. Reasoning about spatial behavior features in the context of the task can facilitate achieving reasonable prediction performance with lower sensor requirements. Also, occlusions are likely more common for facial expressions than body tracking, \track{as we observed in our real-world demonstration}. 

Second, \track{it is important to consider the granularity of  the predictions over \thrirevision{perceptions} of robot performance. We began our work gathering \thrirevision{perceptions} of robot performance on a 5-point Likert responding format, which we believed could reveal subtle aspects of  human perceptions during navigation. However, we found that predicting \thrirevision{perceptions} of robot performance over 5 classes was challenging for both humans and ML models. While human prediction performance could have been affected by specific details of the visualizations that we used to gather our human baseline results, it is worth considering less granular feedback to favor prediction performance during robot deployments. In particular,  for more practical usage of human feedback,}
we recommend building models that \track{start by} identifying poor robot performance (performing binary classification) \track{and then, on top of that, try to predict} more granular \thrirevision{perceptions} of robot performance. 

Finally, if a robot is executing multiple behaviors, we recommend considering whether the robot switched behaviors recently when reasoning about performance predictions. As in our results, predicting performance  recently after a behavior change can be more difficult than before, when the behavior was more consistent.

\vspace{1em}
\section{Limitations and Future Work}
Our work has several limitations that point to interesting future directions.  \track{In particular}, we obtained human baselines for prediction performance, but used only a limited set of feature combinations \track{that described interactions in a single VR environment and two real-world environments}. In the future, it would be interesting to consider a broader set of feature categories \track{in a more diverse range of environments}. For instance, future work could \track{investigate} the value of more detailed human pose features (e.g., \cite{zheng2019deephuman}) \track{across a wider range of scenarios (public plazas or hospitals) where humans may behave differently due to their activity, stress or other  factors.}

\track{Facial expressions and the nuance of human motion are challenging to capture. In our data collection with virtual reality,} \thrirevision{the use of VR could have biased observed nonverbal behavior as well as human perceptions of a robot, given the way humans provide input to the simulation via the VR device, the way the device captures their nonverbal behavior, and the sim-to-real gap.}

\thrirevision{W}e were limited by the features captured by the Vive Pro Eye VR headset, which describe the geometry of the face through blend shapes.
We visualized this data by rendering the features on a virtual avatar head, and this could have affected the perception of subtle human facial expressions.
In the future, it would be interesting to utilize more advanced devices such as the recently released Apple Vision Pro to create other datasets of implicit human feedback. The new Apple device can sense faces in a way that allows rendering higher quality avatars for users, and the data it captures could potentially improve the accuracy and robustness of ML models that predict robot performance.

\track{In the future, inferred performance predictions could be used to adapt robot behavior.
For example, a robot could use binary robot performance predictions as} instantaneous rewards \track{that guide changes in robot behavior to better align what the robot does with human preferences} \cite{knox2009interactively,macglashan2017interactive,cui2021empathic}. \track{When the predictions indicate low robot performance or suggest drastic changes in \thrirevision{perceptions} of the robot's behavior, the robot could also opt for querying users explicitly about its performance to verify the predictions. Perhaps the responses can also be used to improve the prediction model.}

\section{Conclusion} %
\label{sec:conclusion}
This work contributes the \textsc{SEAN TOGETHER} Dataset, consisting of observations of human-robot interactions in VR, including implicit human feedback, and corresponding performance ratings in guided robot navigation tasks. Our analyses \track{with VR data} revealed that facial expressions can help predict \thrirevision{perceptions} of the robot, but spatial behavior features in the context of the navigation task were more critical for these inferences. Our experiments also demonstrated the ability of humans and ML models to infer perceived robot performance from interaction observations. \track{A general trend that we observed throughout this work was that} predicting the directionality of \thrirevision{perceptions} of robot performance (as a binary classification task) \track{was easier and, thus, seemed more practical} than predicting exact performance ratings (on a 5-point scale).

\track{As part of this work, we also conducted a real-world demonstration that showed the applicability of machine learning in predicting human perceptions of a mobile robot in indoor environments. We did not capture facial expression features for this demonstration, but rather focused on capturing features that described the navigation behavior of the robot and humans based on our prior findings. 
} 
\thri{Both the models trained with VR data and real-world data showed promising generalization capabilities when evaluated on real-world data, confirming the potential of machine learning for predicting \thrirevision{perceptions} of robot performance from implicit feedback signals in social robot navigation.}
Our dataset\track{s}, accompanying analyses, and demonstration \track{facilitate future research on more scalable supervision of robot navigation behavior.} Potentially, \track{robots} could use implicit human feedback \track{as supervision} to interactively improve their behavior in the future. \thrirevision{For example, human perceptions of robot performance predicted by machine learning models could be used as a reward function in a reinforcement-learning setup, where the robot improves its policy to learn how to best navigate with a user.}

\section*{Acknowledgments}

\thri{This work was partially supported by the National Science Foundation (Grant No. IIS-1924802, IIS-2143109, and IIS-2106690) and Google. We are grateful to Carolina Parada and Leila Takayama for their valuable feedback.}

\appendix
\section*{Appendix}

\renewcommand{\thefigure}{A\arabic{figure}}

\setcounter{figure}{0}

\renewcommand{\thetable}{A\arabic{table}}

\setcounter{table}{0}


This appendix first shows the distributions of ground truth labels provided by participants over 10-minute intervals of the data collection sessions. \track{Second}, we provide additional details about the annotation tool that we used to collect human annotators’ predictions of users' \thrirevision{perceptions} of robot performance, \track{analyze inter-rater reliability for the human annotations, }and \track{discuss} further findings from the human annotation samples. \track{Then, we provide the full list of features used for predicting human \thrirevision{perceptions} of robot performance, with a brief description of each feature.} Lastly, we describe the specific model architectures that we used and our training procedure. These details are included in this document to facilitate better understanding of our methodology and reproducibility of our work.

\section{Distribution of Ground-Truth Labels Given by Participants over 10-minute Intervals}



Fig.~\ref{fig:histogram_distribution}(a) shows the distributions of ground-truth labels provided by the participants of our VR data collection (Sec. \ref{sec:vr} in the paper) over 10-minute intervals of the data collection sessions. Fig.~\ref{fig:histogram_distribution}(b) shows the distribution of labels from the 120 samples that we randomly drew for human annotation (Sec. \ref{sec:human_procedure} and \ref{sec:analysis-ml-vs-human} in the paper). Overall, the distributions of labels over different intervals are similar, except for 30-40 min, which is close to the end of navigation tasks.

\section{Human Annotation} 

\subsection{\track{Annotation Interface}}

\track{To reduce misalignment between human annotators, we conducted a couple of pilots for our data collection with the team and lab members, through which we improved our annotation interface and instructions. Fig.~\ref{fig:instruction_p1} and \ref{fig:instruction_p2} show the instruction pages in our annotation tool.}


\begin{figure}[t!p]
    \centering
    \includegraphics[width=0.98\linewidth]{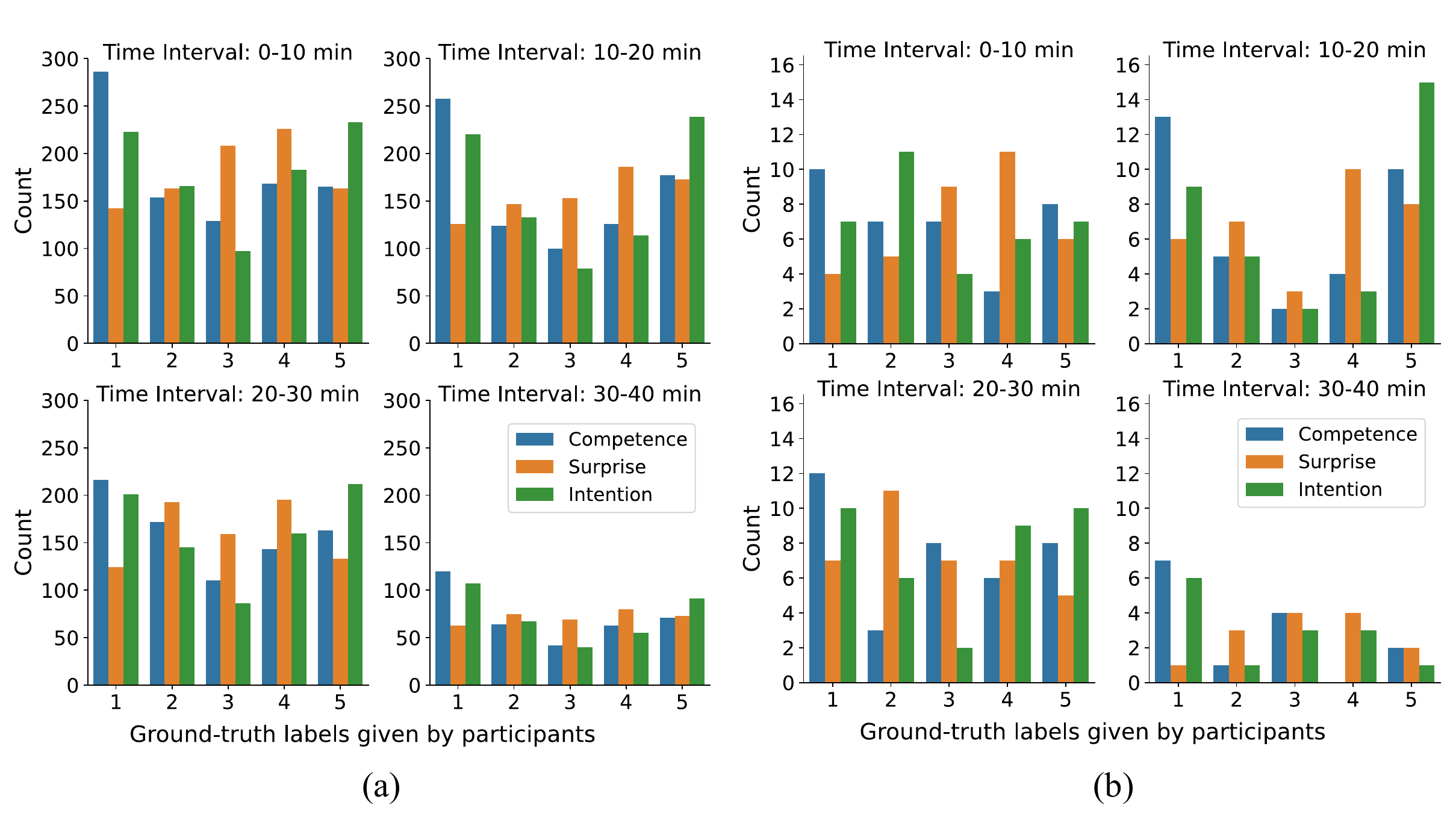}
    \caption{(a) Distributions of ground truth labels provided by the participants that experienced the human-robot interactions in VR over 10-minute intervals of the data collection sessions; (b) Distributions of ground truth labels used for the human annotation. They are a subset of those in (a). Each plot shows data over 10-minute intervals of the data collection. 
    }
    \label{fig:histogram_distribution}
    \vspace{-1em}
\end{figure}

\subsection{\track{Annotation Reliability}}

\track{For each visualization of a data sample, we asked 10 different human annotators to provide their predictions on it,  which allowed us to compare their prediction performance statistically as reported in the paper. 
In addition to those results, we also evaluated the reliability of the human annotations. 
More specifically, we used Krippendorff's alpha to measure the inter-rater reliability for our ordinal labels, which led to an $\alpha$ of 0.67 for competence, 0.54 for surprise, and 0.68 for intention, respectively. These values indicate a moderate to substantial level of agreement among the annotators.
}

\track{Fig. \ref{fig:reliability} shows the distribution of labels given by human annotators on the 120 data samples considered for our human baseline.}

\section{Further Findings from Human Annotation Samples}

Upon reviewing the data that we had collected, we realized that the renderings of participants' faces were shown to annotators with the face mirrored. We were concerned this could have led  to confusion among the  annotators when they evaluated  the gaze direction of the face rendering in comparison to the  navigation rendering (e.g., \track{as in Fig. \ref{fig:prolific} of the paper}). Therefore, we repeated the data collection for the  \textit{Nav.+Facial} condition, but with the face image not mirrored. 

Results for the mirror and not-mirrored data are shown in Table \ref{tbl:mirror_results}. The results in the not-mirrored case only had a subtle difference in comparison to the mirrored case. This suggest that the gaze direction was not an issue and validate the reproducibility of our human annotation experiments.

As discussed in Sec. \ref{sec:analysis-ml-vs-human} of the paper, for the 120 data samples shown to the human annotators, the performance of machine learning models for predicting Intention with \textit{Nav.} and \textit{Nav.+Facial} features decreased in the last two time intervals of data collection. As suggested by Fig.~\ref{fig:histogram_distribution}(b), a change in the distribution of ground truth labels can be observed in the interval of 30-40 min. This interval also had the fewest number of samples due to how the data was collected, because not all interactions took the full 40 minutes. Also, a good proportion of the samples in the last time interval showed the end of navigation tasks, at which point the participants could have been more sensitive to robot navigation in the wrong direction. Indeed, there was a higher proportion of lower ratings for Intention in the last interval than in the other intervals. Such a distribution shift can make the prediction task harder for both the annotators and the machine learning models.

\begin{figure}[t!p]
    \centering
    \includegraphics[width=0.48\linewidth]{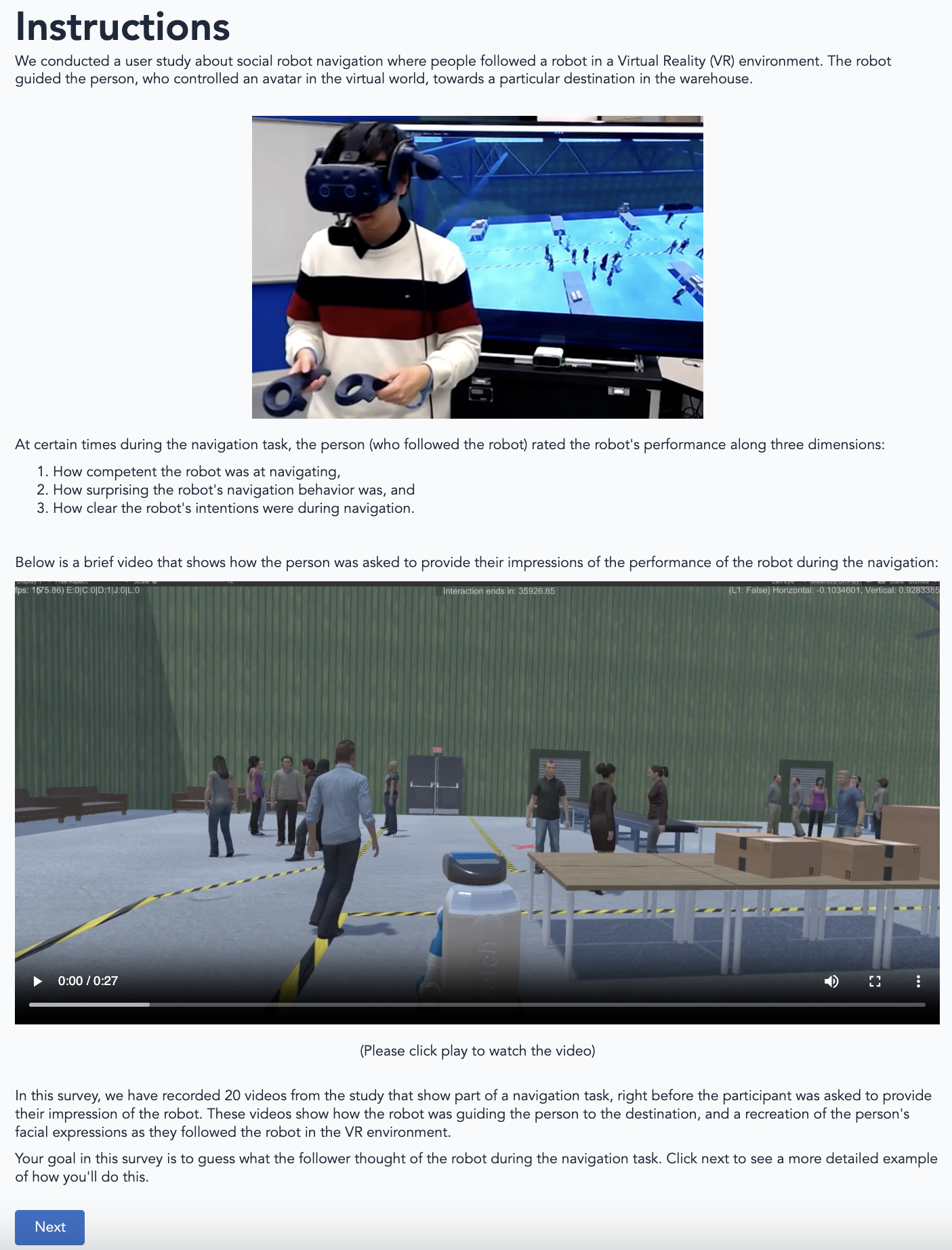}
    \caption{\track{Instruction page 1 that provides the background of social navigation data collected in VR.} 
    }
    \label{fig:instruction_p1}
    \vspace{1.5em}
\end{figure}
\begin{figure}[b!p]
    \includegraphics[fbox=0.5pt 20pt,width=17em]{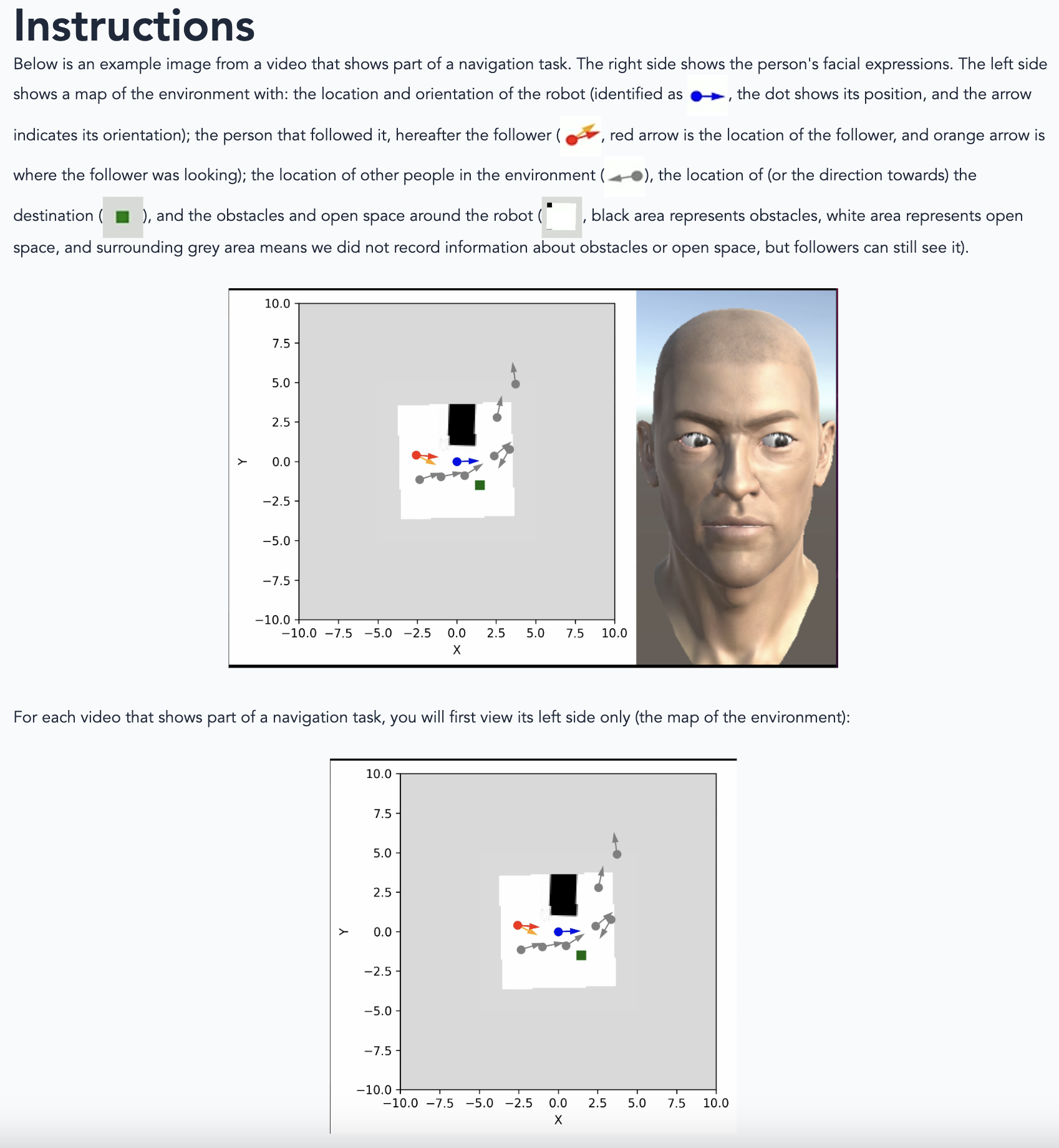}\hspace{1em}
    \includegraphics[fbox=0.5pt 20pt,width=17em]{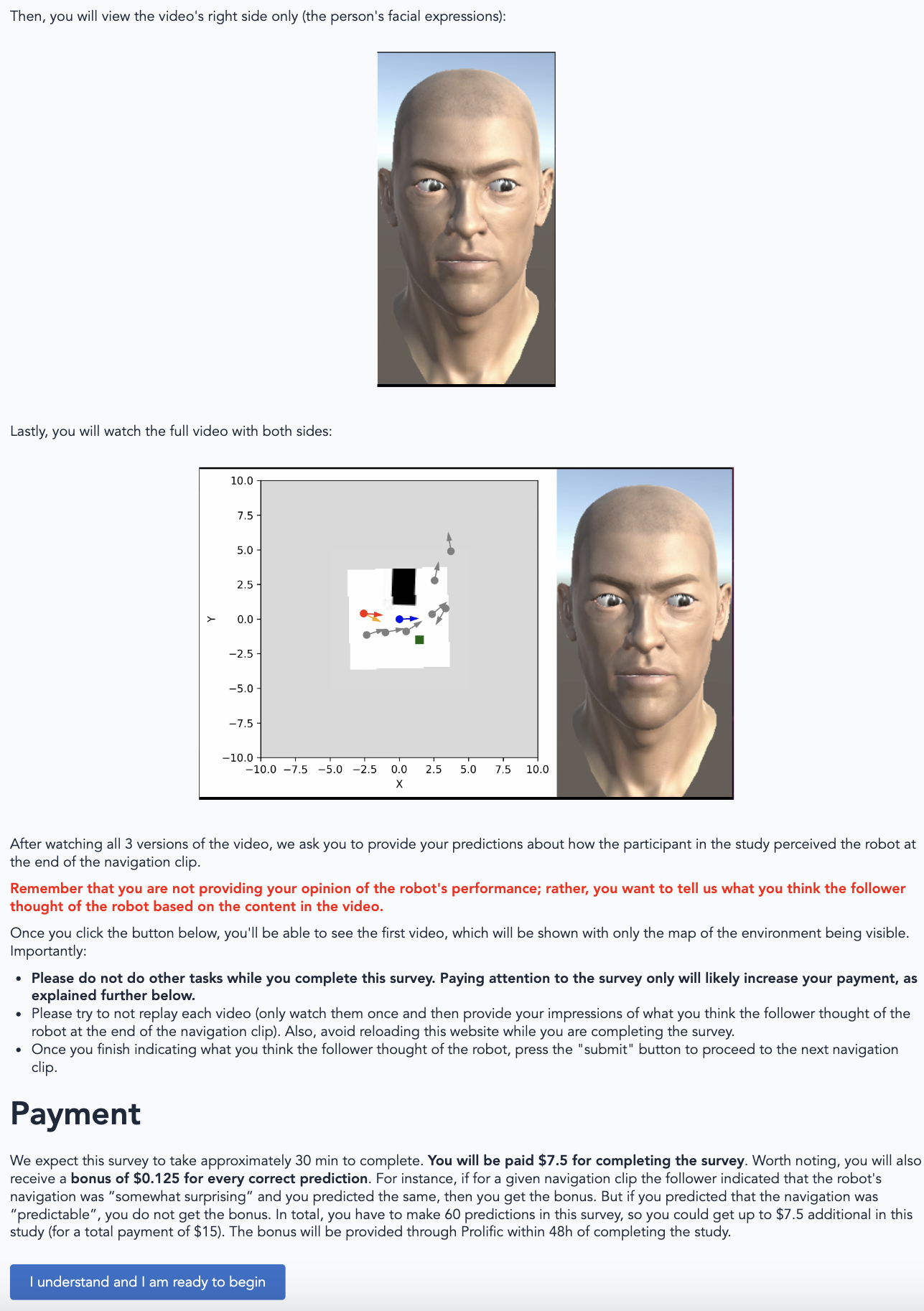}
    \caption{\track{Instruction page 2 that details the annotation procedures and participant's compensation. The left image is the top of the page while the right image is the continuation of the left image.}
    }
    \label{fig:instruction_p2}
\end{figure}

\begin{figure}[bt!p]
    \centering
    \includegraphics[width=0.4\linewidth]{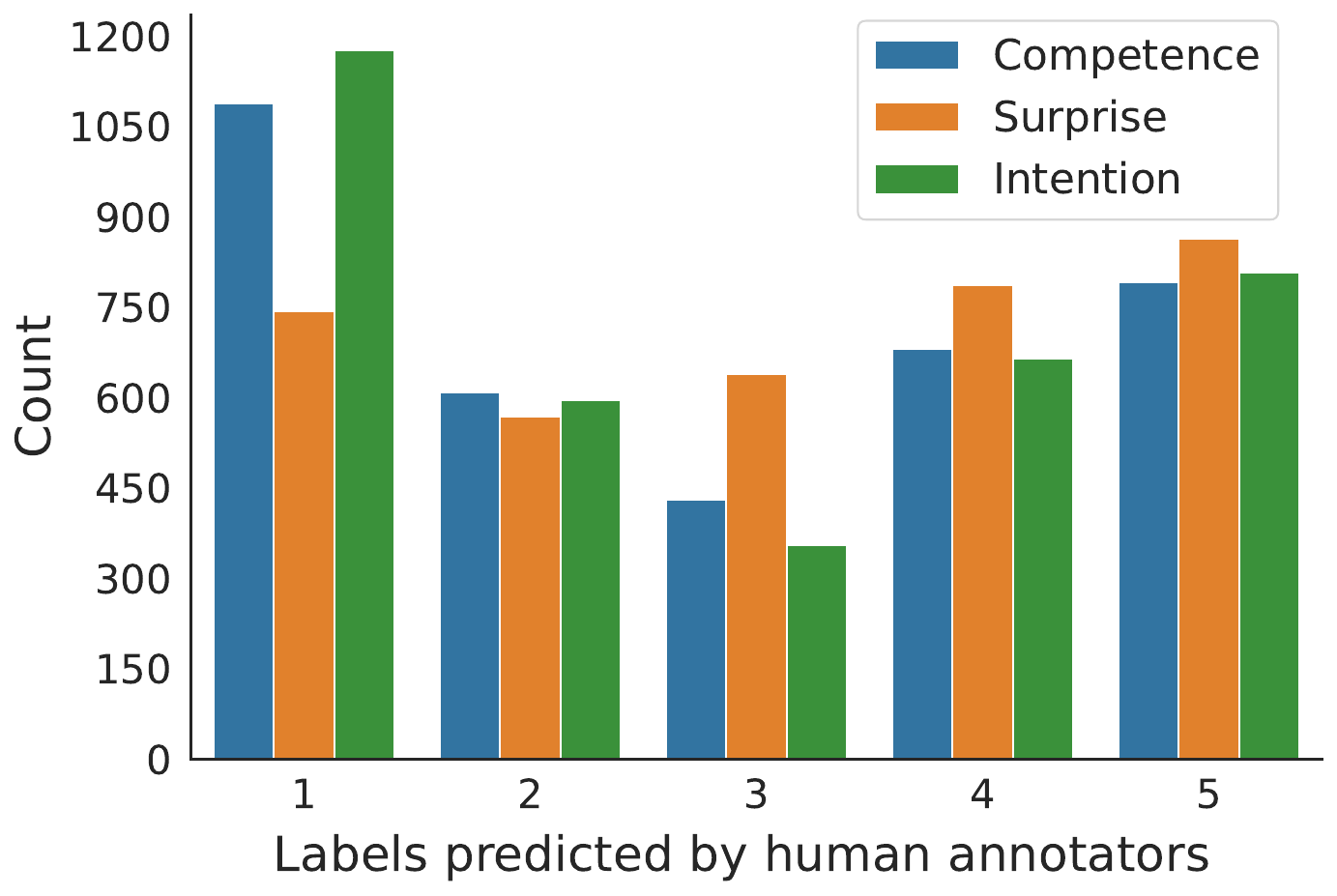}
    \caption{\track{Distribution of labels predicted by 
human annotators.}
    }
    \label{fig:reliability}
\end{figure}

\begin{table}[t!p]
  \caption{The $F_1$ Score, the Accuracy, and the Mean Absolute Error of using mirrored and not-mirrored facial rendering when annotating with \textit{Nav.+Facial} features.
  }
  \label{tbl:mirror_results}
  \centering
{

\begin{adjustbox}{width=0.9\textwidth}
\begin{tabular}{rcccccc}

& \multicolumn{2}{c}{$F_1$-Score ($\mu \pm \sigma$) $\uparrow$}  & \multicolumn{2}{c}{Accuracy ($\mu \pm \sigma$) $\uparrow$}  & \multicolumn{2}{c}{Mean Absolute Error ($\mu \pm \sigma$) $\downarrow$}  \\
\cmidrule(r){2-7}
 & Mirrored & Not-Mirrored & Mirrored & Not-Mirrored & Mirrored & Not-Mirrored \\
\cmidrule(r){1-7} 
Competence &  $0.30 \pm 0.1$   &   $0.29 \pm 0.2$ &  $0.43 \pm 0.1$   &  $0.42 \pm 0.1$ &  $0.96 \pm 0.3$  &  $0.99 \pm 0.4$  \\ 
\cmidrule(r){1-7}
Surprise &  $0.25 \pm 0.1$ &  $0.25 \pm 0.1$ & $0.33 \pm 0.1$  &  $0.32 \pm 0.1$  &  $1.09 \pm 0.2$  &  $1.12 \pm 0.2$  \\ 
\cmidrule(r){1-7}
Intention &  $0.30 \pm 0.1$  &  $0.28 \pm 0.1$  &   $0.42 \pm 0.1$ &  ${0.41 \pm 0.1}$  &  $1.04 \pm 0.3$  &  ${1.07 \pm 0.2}$ \\ 

\end{tabular}
\end{adjustbox}
}
\end{table}



\vspace{-1em}
\track{\section{Feature Extraction}}

\track{The following sections describe in detail the features used for predicting human \thrirevision{perceptions} of robot performance:}
\vspace{0.5em}

\track{
\textbf{Participants’ Facial Expression Features:}
\begin{itemize}
    \item \textit{gaze\_origin\_mm\_[left, right]\_[x, y, z]}: The gaze origins of left and right eyes, measured in millimeters.
    \item \textit{gaze\_direction\_normalized\_[left, right]\_[x, y, z]}: The normalized gaze directions of left and right eyes. 
    \item \textit{pupil\_diameter\_mm\_[left, right]}: The pupil diameters of left and right eyes, measured in millimeters.
    \item \textit{eye\_openness\_[left, right]}: The openness of left and right eyes.
    \item \textit{pupil\_position\_in\_sensor\_area\_[left, right]\_[x, y]}: The pupil positions of left and right eyes in the sensor area.
    \item \textit{gaze\_origin\_mm\_combined\_[x, y, z]}: The combined gaze origin of left and right eyes, measured in millimeters.
    \item \textit{gaze\_direction\_normalized\_combined\_[x, y, z]}: The normalized combined gaze direction of left and right eyes.
    \item \textit{pupil\_diameter\_mm\_combined}: The combined pupil diameter of the left and right eyes, measured in millimeters.
    \item \textit{eye\_openness\_combined}: The combined eye openness of the left and right eyes.
    \item \textit{pupil\_position\_in\_sensor\_area\_combined\_[x, y]}: The combined pupil position of left and right eyes in the sensor area.
    \item \textit{eye\_[wide, squeeze, frown]\_[left, right]}: The extent of eye wide, squeeze, and frown of left and right eyes.
    \item \textit{jaw\_[right, left, forward, open]}: The extent of jaw being right, left, forward, and open.
    \item \textit{mouth\_ape\_shape}: The extent of mouth ape shape.
    \item \textit{mouth\_[upper, lower]\_[right, left]}: The extent of upper and lower part of mouth moving to the right and left.
    \item \textit{mouth\_[upper, lower]\_overturn}: The extent of upper and lower part of mouth overturning.
    \item \textit{mouth\_pout} The extent of mouth pouting.
    \item \textit{mouth\_smile\_[right, left]}: The extent of mouth smiling on the right and left side.
    \item \textit{mouth\_sad\_[right, left]}: The extent of mouth being sad on the right and left side.
    \item \textit{cheek\_puff\_[right, left]}: The extent of cheek puffing on the right and left side.
    \item \textit{cheek\_suck}: The extent of cheek sucking on the right and left side.
    \item \textit{mouth\_upper\_[upright, upleft]}: The extent of the upper part of mouth moving upright and upleft.
    \item \textit{mouth\_lower\_[downright, downleft]}: The extent of the lower part of mouth moving downright and downleft.
    \item \textit{mouth\_[upper, lower]\_inside}: The extent of the upper and lower part of mouth moving inside.
    \item \textit{mouth\_lower\_overlay}: The extent of the lower part of mouth overlaying.
    \item \textit{tongue\_[longstep1, longstep2]}: The extent of the person's tongue stretching long.
    \item \textit{tongue\_[down, up, right, left, roll]}: The extent of the person's tongue moving down, up, right, left, and rolling. 
    \item \textit{tongue\_[upleft, upright, downleft, downright]\_morph}: The extent of the person's  tongue morphing upleft, upright, downleft, and downright.
\end{itemize}}

\track{
\textbf{Spatial Behavior Features:}
\begin{itemize}
    \item \textit{participant\_pose\_[x, y, cos($\theta$), sin($\theta$)]}: The 2D position and orientation of the participant, computed relative to the robot.
    \item \textit{nearby\_agents\_pose\_[x, y, cos($\theta$), sin($\theta$)]}: The 2D positions and orientations of the other automatically-controlled avatars within a 7.2m radius, computed relative to the robot.
\end{itemize}}

\track{
\textbf{Goal Features:}
\begin{itemize}
    \item \textit{goal\_[x, y]}: The 2D position of the navigation destination in a coordinate frame attached to the robot.
\end{itemize}
}

\track{
\textbf{Occupancy Features:}
\begin{itemize}
    \item \textit{map\_resnet18}: The cropped section of the 2D map around the robot (of $7.2$m $\times$ $7.2$m) to describe the occupancy of nearby space by static objects, encoded by ResNet-18 \cite{he2016deep}.
\end{itemize}}

\section{Model Architecture and Hyperparameters}

For training our machine learning models, the input for each example corresponded to an 8-second window of features (navigation, facial, or both types of features). The synchronized multi-modal data was re-sampled at 5Hz, resulting in an input sequence of 40 timesteps. This helped reduce the length of the series of features input to the model, facilitating learning in practice.  The targets for the examples corresponded to the ground truth labels for robot performance at the end of the window. These ground truth labels were provided by the participants in our VR data collection.

For part of our evaluation, we converted the 5-point ratings in the ground truth labels to binary ratings (e.g., as reported in Sec. \ref{sec:human_procedure} of the paper) as well as converted the output of machine learning models and human ratings from a 5-point scale to a binary output. This conversion was used to evaluate how well human annotators and machine learning models could predict the directionality of robot performance (rather than focusing on the exact performance level indicated in the ground truth labels). The binary classification task could be useful in the future for identifying situations were a robot makes mistakes during navigation and potentially engaging in recovery behaviors.

To facilitate future reproducibility, the next paragraphs provide more details about the specific  architectures implemented for the deep learning models considered in our work:
\vspace{0.5em}

\noindent
\textbf{MLP architecture.} Our MLP model first encoded the input features at each timestep with a dense linear layer with 256 hidden units. The encoded sequence was then concatenated across timesteps and fed into three nonlinear dense layers with 512, 256, and 64 hidden units, respectively, and with Leaky ReLU activation. Finally, it was passed into a linear layer that output the logits corresponding to the 5 categories of labels to be classified.
\vspace{0.5em}

\noindent
\textbf{Transformer architecture.} 
Our Transformer model first passed the input sequence through a BatchNorm layer, and then encoded at each timestep with a dense layer with 256 hidden units. Positional encoding was applied to the encoded sequence, which was then fed into 2 transformer encoder layers with 4 heads, a feed-forward dimension of 512, and ReLU activation. The output was then concatenated across timesteps and fed into three nonlinear dense layers with 512, 256, and 64 hidden units, respectively, and with Leaky ReLU activation. Finally, the result was passed into a linear layer that output the logits corresponding to the 5 categories of labels to be classified.
\vspace{0.5em}

\noindent
\textbf{Graph Neural Network (GNN) architecture.} We constructed a bidirectional, fully-connected graph in order to utilize the relational inductive bias present in the data and process this data using a GNN.
Input sequences of temporal data were first divided into three groups, corresponding to the node features, the edge features, and the global features.
Node features consisted of the positions and orientations of the participant and nearby agents relative to the robot.
Edge features between every pair of nodes in the graph consisted of the Euclidean distance between the two connected nodes. 
Global features consisted of all other features for a given experiment.
A feedforward network with 64 hidden units was created for each of the three groups of temporal data.
Then, each time step of each type of temporal data was encoded using the corresponding feedforward network.

The architecture of our model consisted of two message-passing layers~\cite{battaglia2018relational}.
Each edge update function and each node update function was composed of a feedforward network with a ReLU activated, single hidden layer of 64 units.
All of the node representations for a graph that resulted from the final message-passing layer were concatenated with the global feature representations that resulted from the temporal encoding of the input global features.
Finally, a classification head of three, Leakly ReLU activated, nonlinear dense layers with 512, 256, and 64 hidden units, respectively, was used to output the logits corresponding to the 5 categories of labels to be classified. 
\vspace{0.5em}

\begin{table}[tb]
  \caption{The best learning rate, batch size, and dropout of Multi-Layer Perceptron (MLP), Graph Neural Network (GNN), and Transformer (T), chosen using grid search with validation-based early stopping. ``Annotation Samples'' values correspond to the hyper-parameters for the deep learning models reported in Table I of the paper (Sec. 5.2). ``Leave-One-Out'' values correspond to the hyper-parameters for the results in Fig. 6 of the paper (Sec. 5.3).   
  }
  \label{tbl:hyperparameters}
  \centering
{

\begin{adjustbox}{width=0.95\textwidth}
\begin{tabular}{rcccccc}

& \multicolumn{2}{c}{Learning Rate}  & \multicolumn{2}{c}{Batch Size}  & \multicolumn{2}{c}{Dropout}  \\
\cmidrule(r){2-7}
 & Annotation Samples & Leave-One-Out & Annotation Samples & Leave-One-Out & Annotation Samples & Leave-One-Out \\
\cmidrule(r){1-7} 
MLP &  $0.003$  &  $0.001$  &  $256$  &  $512$  &  $0.1$   &  $0.1$  \\ 
\cmidrule(r){1-7}
GNN &  $0.003$  &  $0.001$  &  $512$  &  $512$  &  $0.0$   &  $0.5$  \\ 
\cmidrule(r){1-7}
T &  $0.003$  &  $0.0003$  &  $512$  &  $512$  &  $0.0$   &  $0.0$  \\ 

\end{tabular}
\end{adjustbox}
}
\vspace{-1em}
\end{table}

The deep learning models were trained using the Cross-Entropy (CE) loss against the ground truth labels. To update model parameters, we used the AdamW optimizer with a weight decay coefficient of 0.01. The best learning rates, batch sizes, and dropout rates found by hyperparameter search are shown in Table~\ref{tbl:hyperparameters}. All the results described in the paper were obtained with an Intel Core i7 10700K 8-Core 3.6GHz desktop computer that had an NVIDIA 24GB GeForce RTX 3090 GPU.





\bibliographystyle{ACM-Reference-Format}
\bibliography{references}

@inproceedings{cui2021empathic,
  title={The empathic framework for task learning from implicit human feedback},
  author={Cui, Yuchen and Zhang, Qiping and Knox, Brad and Allievi, Alessandro and Stone, Peter and Niekum, Scott},
  booktitle={Conference on Robot Learning},
  pages={604--626},
  year={2021},
  organization={PMLR}
}

@inproceedings{zhang2023self,
  title={Self-Annotation Methods for Aligning Implicit and Explicit Human Feedback in Human-Robot Interaction},
  author={Zhang, Qiping and Narcomey, Austin and Candon, Kate and V{\'a}zquez, Marynel},
  booktitle={Proceedings of the 2023 ACM/IEEE International Conference on Human-Robot Interaction},
  pages={398--407},
  year={2023}
}

@inproceedings{aronson2018gaze,
  title={Gaze for error detection during human-robot shared manipulation},
  author={Aronson, Reuben M and Admoni, Henny},
  booktitle={Fundamentals of Joint Action workshop, Robotics: Science and Systems},
  pages={5},
  year={2018}
}

@article{kendon1988goffman,
  title={Goffman's approach to face-to-face interaction},
  author={Kendon, Adam},
  journal={Erving Goffman: Exploring the interaction order},
  year={1988},
  publisher={Polity Press}
}

@inproceedings{sadigh2016planning,
  title={Planning for autonomous cars that leverage effects on human actions.},
  author={Sadigh, Dorsa and Sastry, Shankar and Seshia, Sanjit A and Dragan, Anca D},
  booktitle={Robotics: Science and systems},
  volume={2},
  pages={1--9},
  year={2016},
  organization={Ann Arbor, MI, USA}
}

@inproceedings{zhang2023sean,
  title={SEAN-VR: An Immersive Virtual Reality Experience for Evaluating Social Robot Navigation},
  author={Zhang, Qiping and Tsoi, Nathan and V{\'a}zquez, Marynel},
  booktitle={Companion of the 2023 ACM/IEEE International Conference on Human-Robot Interaction},
  pages={902--904},
  year={2023}
}

@inproceedings{asavanant2021personal,
  title={Personal space violation by a robot: An application of expectation violation theory in human-robot interaction},
  author={Asavanant, Chatchalita and Umemuro, Hiroyuki},
  booktitle={2021 30th IEEE International Conference on Robot \& Human Interactive Communication (RO-MAN)},
  pages={1181--1188},
  year={2021},
  organization={IEEE}
}

@article{vaswani2017attention,
  title={Attention is all you need},
  author={Vaswani, Ashish and Shazeer, Noam and Parmar, Niki and Uszkoreit, Jakob and Jones, Llion and Gomez, Aidan N and Kaiser, {\L}ukasz and Polosukhin, Illia},
  journal={Advances in neural information processing systems},
  volume={30},
  year={2017}
}

@article{battaglia2018relational,
  title={Relational inductive biases, deep learning, and graph networks},
  author={Battaglia, Peter W and Hamrick, Jessica B and Bapst, Victor and Sanchez-Gonzalez, Alvaro and Zambaldi, Vinicius and Malinowski, Mateusz and Tacchetti, Andrea and Raposo, David and Santoro, Adam and Faulkner, Ryan and others},
  journal={arXiv preprint arXiv:1806.01261},
  year={2018}
}

@inproceedings{bera2019improving,
  title={Improving Socially-aware Multi-channel Human Emotion Prediction for Robot Navigation.},
  author={Bera, Aniket and Randhavane, Tanmay and Manocha, Dinesh},
  booktitle={CVPR Workshops},
  pages={21--27},
  year={2019}
}

@article{francis2023principles,
  title={Principles and guidelines for evaluating social robot navigation algorithms},
  author={Francis, Anthony and P{\'e}rez-d'Arpino, Claudia and Li, Chengshu and Xia, Fei and Alahi, Alexandre and Alami, Rachid and Bera, Aniket and Biswas, Abhijat and Biswas, Joydeep and Chandra, Rohan and others},
  journal={arXiv preprint arXiv:2306.16740},
  year={2023}
}

@article{gao2022evaluation,
  title={Evaluation of socially-aware robot navigation},
  author={Gao, Yuxiang and Huang, Chien-Ming},
  journal={Frontiers in Robotics and AI},
  volume={8},
  pages={721317},
  year={2022},
  publisher={Frontiers}
}

@article{mavrogiannis2023core,
  title={Core challenges of social robot navigation: A survey},
  author={Mavrogiannis, Christoforos and Baldini, Francesca and Wang, Allan and Zhao, Dapeng and Trautman, Pete and Steinfeld, Aaron and Oh, Jean},
  journal={ACM Transactions on Human-Robot Interaction},
  volume={12},
  number={3},
  pages={1--39},
  year={2023},
  publisher={ACM New York, NY}
}

@inproceedings{stiber2022modeling,
  title={Modeling Human Response to Robot Errors for Timely Error Detection},
  author={Stiber, Maia and Taylor, Russell and Huang, Chien-Ming},
  booktitle={2022 IEEE/RSJ International Conference on Intelligent Robots and Systems (IROS)},
  pages={676--683},
  year={2022},
  organization={IEEE}
}

@inproceedings{wachowiak2022analysing,
  title={Analysing eye gaze patterns during confusion and errors in human--agent collaborations},
  author={Wachowiak, Lennart and Tisnikar, Peter and Canal, Gerard and Coles, Andrew and Leonetti, Matteo and Celiktutan, Oya},
  booktitle={2022 31st IEEE International Conference on Robot and Human Interactive Communication (RO-MAN)},
  pages={224--229},
  year={2022},
  organization={IEEE}
}

@inproceedings{stiber2022effective,
  title={Effective Human-Robot Collaboration via Generalized Robot Error Management Using Natural Human Responses},
  author={Stiber, Maia},
  booktitle={Proceedings of the 2022 International Conference on Multimodal Interaction},
  pages={673--678},
  year={2022}
}

@inproceedings{angelopoulos2022you,
  title={You Are In My Way: Non-verbal Social Cues for Legible Robot Navigation Behaviors},
  author={Angelopoulos, Georgios and Rossi, Alessandra and Di Napoli, Claudia and Rossi, Silvia},
  booktitle={2022 IEEE/RSJ International Conference on Intelligent Robots and Systems (IROS)},
  pages={657--662},
  year={2022},
  organization={IEEE}
}

@inproceedings{tsoi2021approach,
  title={An approach to deploy interactive robotic simulators on the web for hri experiments: Results in social robot navigation},
  author={Tsoi, Nathan and Hussein, Mohamed and Fugikawa, Olivia and Zhao, JD and V{\'a}zquez, Marynel},
  booktitle={2021 IEEE/RSJ International Conference on Intelligent Robots and Systems (IROS)},
  pages={7528--7535},
  year={2021},
  organization={IEEE}
}

@inproceedings{mcquillin2022learning,
  title={Learning socially appropriate robo-waiter behaviours through real-time user feedback},
  author={McQuillin, Emily and Churamani, Nikhil and Gunes, Hatice},
  booktitle={2022 17th ACM/IEEE International Conference on Human-Robot Interaction (HRI)},
  pages={541--550},
  year={2022},
  organization={IEEE}
}

@inproceedings{yan2020frownonerror,
  title={Frownonerror: Interrupting responses from smart speakers by facial expressions},
  author={Yan, Yukang and Yu, Chun and Zheng, Wengrui and Tang, Ruining and Xu, Xuhai and Shi, Yuanchun},
  booktitle={Proceedings of the 2020 CHI Conference on Human Factors in Computing Systems},
  pages={1--14},
  year={2020}
}

@article{lin2020review,
  title={A review on interactive reinforcement learning from human social feedback},
  author={Lin, Jinying and Ma, Zhen and Gomez, Randy and Nakamura, Keisuke and He, Bo and Li, Guangliang},
  journal={IEEE Access},
  volume={8},
  pages={120757--120765},
  year={2020},
  publisher={IEEE}
}

@inproceedings{carpinella2017robotic,
  title={The robotic social attributes scale (rosas) development and validation},
  author={Carpinella, Colleen M and Wyman, Alisa B and Perez, Michael A and Stroessner, Steven J},
  booktitle={Proceedings of the 2017 ACM/IEEE International Conference on human-robot interaction},
  pages={254--262},
  year={2017}
}

@inproceedings{candon2023nonverbal,
  title={Nonverbal Human Signals Can Help Autonomous Agents Infer Human Preferences for Their Behavior},
  author={Candon, Kate and Chen, Jesse and Kim, Yoony and Hsu, Zoe and Tsoi, Nathan and and V{\'a}zquez, Marynel},
  booktitle={Proceedings of the 22nd International Conference on Autonomous Agents and Multiagent Systems},
  year={2023}
}

@article{rivoire2016delicate,
  title={The delicate balance of boring and annoying: Learning proactive timing in long-term human robot interaction},
  author={Rivoire, Claire and Lim, Angelica},
  year={2016}
}

@inproceedings{kidokoro2013will,
  title={Will I bother here?-A robot anticipating its influence on pedestrian walking comfort},
  author={Kidokoro, Hiroyuki and Kanda, Takayuki and Br{\v{s}}cic, Dra{\v{z}}en and Shiomi, Masahiro},
  booktitle={2013 8th ACM/IEEE International Conference on Human-Robot Interaction (HRI)},
  pages={259--266},
  year={2013},
  organization={IEEE}
}

@article{rubagotti2022perceived,
  title={Perceived safety in physical human--robot interaction—A survey},
  author={Rubagotti, Matteo and Tusseyeva, Inara and Baltabayeva, Sara and Summers, Danna and Sandygulova, Anara},
  journal={Robotics and Autonomous Systems},
  volume={151},
  pages={104047},
  year={2022},
  publisher={Elsevier}
}

@article{akalin2022you,
  title={Do you feel safe with your robot? Factors influencing perceived safety in human-robot interaction based on subjective and objective measures},
  author={Akalin, Neziha and Kristoffersson, Annica and Loutfi, Amy},
  journal={International journal of human-computer studies},
  volume={158},
  pages={102744},
  year={2022},
  publisher={Elsevier}
}

@inproceedings{gucsi2020ask,
  title={To ask or not to ask: A user annoyance aware preference elicitation framework for social robots},
  author={Gucsi, Balint and Tarapore, Danesh S and Yeoh, William and Amato, Christopher and Tran-Thanh, Long},
  booktitle={2020 IEEE/RSJ International Conference on Intelligent Robots and Systems (IROS)},
  pages={7935--7940},
  year={2020},
  organization={IEEE}
}

@article{tsoi2022sean,
  title={Sean 2.0: Formalizing and generating social situations for robot navigation},
  author={Tsoi, Nathan and Xiang, Alec and Yu, Peter and Sohn, Samuel S and Schwartz, Greg and Ramesh, Subashri and Hussein, Mohamed and Gupta, Anjali W and Kapadia, Mubbasir and V{\'a}zquez, Marynel},
  journal={IEEE Robotics and Automation Letters},
  volume={7},
  number={4},
  pages={11047--11054},
  year={2022},
  publisher={IEEE}
}

@inproceedings{brandao2021experts,
  title={How experts explain motion planner output: a preliminary user-study to inform the design of explainable planners},
  author={Brandao, Martim and Canal, Gerard and Krivi{\'c}, Senka and Luff, Paul and Coles, Amanda},
  booktitle={2021 30th IEEE International Conference on Robot \& Human Interactive Communication (RO-MAN)},
  pages={299--306},
  year={2021},
  organization={IEEE}
}

@inproceedings{dragan2013legibility,
  title={Legibility and predictability of robot motion},
  author={Dragan, Anca D and Lee, Kenton CT and Srinivasa, Siddhartha S},
  booktitle={2013 8th ACM/IEEE International Conference on Human-Robot Interaction (HRI)},
  pages={301--308},
  year={2013},
  organization={IEEE}
}

@book{hall1966hidden,
  title={The hidden dimension},
  author={Hall, Edmund T and Hall, Edward T},
  volume={609},
  year={1966},
  publisher={Anchor}
}

@article{sciutti2018humanizing,
  title={Humanizing human-robot interaction: On the importance of mutual understanding},
  author={Sciutti, Alessandra and Mara, Martina and Tagliasco, Vincenzo and Sandini, Giulio},
  journal={IEEE Technology and Society Magazine},
  volume={37},
  number={1},
  pages={22--29},
  year={2018},
  publisher={IEEE}
}

@inproceedings{dragan2015effects,
  title={Effects of robot motion on human-robot collaboration},
  author={Dragan, Anca D and Bauman, Shira and Forlizzi, Jodi and Srinivasa, Siddhartha S},
  booktitle={Proceedings of the Tenth Annual ACM/IEEE International Conference on Human-Robot Interaction},
  pages={51--58},
  year={2015}
}

@incollection{gunes2008lab,
  title={From the Lab to the Real World: Affect Recognition Using Multiple Cues and Modalities},
  author={Gunes, Hatice and Piccardi, Massimo and Pantic, Maja},
  booktitle={Affective Computing},
  year={2008},
  publisher={IntechOpen}
}

@article{vinciarelli2009social,
  title={Social signal processing: Survey of an emerging domain},
  author={Vinciarelli, Alessandro and Pantic, Maja and Bourlard, Herv{\'e}},
  journal={Image and vision computing},
  volume={27},
  number={12},
  pages={1743--1759},
  year={2009},
  publisher={Elsevier}
}

@inproceedings{stiber23hri,
author = {Stiber, Maia and Taylor, Russell H. and Huang, Chien-Ming},
title = {On Using Social Signals to Enable Flexible Error-Aware HRI},
year = {2023},
publisher = {Association for Computing Machinery},
address = {New York, NY, USA},
url = {https://doi.org/10.1145/3568162.3576990},
booktitle = {Proceedings of the 2023 ACM/IEEE International Conference on Human-Robot Interaction},
pages = {222–230},
numpages = {9},
location = {Stockholm, Sweden},
series = {HRI '23}
}

@inproceedings{knepper2017implicit,
  title={Implicit communication in a joint action},
  author={Knepper, Ross A and Mavrogiannis, Christoforos I and Proft, Julia and Liang, Claire},
  booktitle={Proceedings of the 2017 acm/ieee international conference on human-robot interaction},
  pages={283--292},
  year={2017}
}

@inproceedings{quigley2009ros,
author = {Quigley, Morgan and Conley, Ken and Gerkey, Brian and Faust, Josh and Foote, Tully and Leibs, Jeremy and Wheeler, Rob and Ng, Andrew},
year = {2009},
month = {01},
pages = {},
title = {ROS: an open-source Robot Operating System},
volume = {3},
journal = {ICRA Workshop on Open Source Software}
}

@article{patterson1975maximum,
 ISSN = {01621459},
 URL = {http://www.jstor.org/stable/2286796},
 author = {David A. Harville},
 journal = {Journal of the American Statistical Association},
 number = {358},
 pages = {320--338},
 publisher = {[American Statistical Association, Taylor & Francis, Ltd.]},
 title = {Maximum Likelihood Approaches to Variance Component Estimation and to Related Problems},
 urldate = {2023-01-02},
 volume = {72},
 year = {1977}
}

@book{stroup2012generalized,
  title={Generalized linear mixed models: modern concepts, methods and applications},
  author={Stroup, Walter W},
  year={2012},
  publisher={CRC press}
}

@inproceedings{lu2014layered,
  title={Layered costmaps for context-sensitive navigation},
  author={Lu, David V and Hershberger, Dave and Smart, William D},
  booktitle={2014 IEEE/RSJ International Conference on Intelligent Robots and Systems},
  pages={709--715},
  year={2014},
  organization={IEEE}
}

@article{grisetti2007improved,
  title={Improved techniques for grid mapping with rao-blackwellized particle filters},
  author={Grisetti, Giorgio and Stachniss, Cyrill and Burgard, Wolfram},
  journal={IEEE transactions on Robotics},
  volume={23},
  number={1},
  pages={34--46},
  year={2007},
  publisher={IEEE}
}

@inproceedings{tan2019one,
  title={From one to another: how robot-robot interaction affects users' perceptions following a transition between robots},
  author={Tan, Xiang Zhi and Reig, Samantha and Carter, Elizabeth J and Steinfeld, Aaron},
  booktitle={2019 14th ACM/IEEE International Conference on Human-Robot Interaction (HRI)},
  pages={114--122},
  year={2019},
  organization={IEEE}
}

@inproceedings{avrunin2014socially,
  title={Socially-appropriate approach paths using human data},
  author={Avrunin, Eleanor and Simmons, Reid},
  booktitle={The 23rd IEEE International Symposium on Robot and Human Interactive Communication},
  pages={1037--1042},
  year={2014},
  organization={IEEE}
}

@article{mavrogiannis2022social,
  title={Social momentum: Design and evaluation of a framework for socially competent robot navigation},
  author={Mavrogiannis, Christoforos and Alves-Oliveira, Patr{\'\i}cia and Thomason, Wil and Knepper, Ross A},
  journal={ACM Transactions on Human-Robot Interaction (THRI)},
  volume={11},
  number={2},
  pages={1--37},
  year={2022},
  publisher={ACM New York, NY}
}

@article{suresh2023robot,
  title={Robot Navigation in Risky, Crowded Environments: Understanding Human Preferences},
  author={Suresh, Aamodh and Taylor, Angelique and Riek, Laurel D and Martinez, Sonia},
  journal={arXiv preprint arXiv:2303.08284},
  year={2023}
}

@inproceedings{biyik2022aprel,
  title={Aprel: A library for active preference-based reward learning algorithms},
  author={B{\i}y{\i}k, Erdem and Talati, Aditi and Sadigh, Dorsa},
  booktitle={2022 17th ACM/IEEE International Conference on Human-Robot Interaction (HRI)},
  pages={613--617},
  year={2022},
  organization={IEEE}
}

@article{thomaz2008teachable,
  title={Teachable robots: Understanding human teaching behavior to build more effective robot learners},
  author={Thomaz, Andrea L and Breazeal, Cynthia},
  journal={Artificial Intelligence},
  volume={172},
  number={6-7},
  pages={716--737},
  year={2008},
  publisher={Elsevier}
}

@inproceedings{lo2019perception,
  title={Perception of pedestrian avoidance strategies of a self-balancing mobile robot},
  author={Lo, Shih-Yun and Yamane, Katsu and Sugiyama, Ken-ichiro},
  booktitle={2019 IEEE/RSJ International Conference on Intelligent Robots and Systems (IROS)},
  pages={1243--1250},
  year={2019},
  organization={IEEE}
}

@inproceedings{cui2021understanding,
  title={Understanding the relationship between interactions and outcomes in human-in-the-loop machine learning},
  author={Cui, Yuchen and Koppol, Pallavi and Admoni, Henny and Niekum, Scott and Simmons, Reid and Steinfeld, Aaron and Fitzgerald, Tesca},
  booktitle={International Joint Conference on Artificial Intelligence},
  year={2021}
}

@article{mitsunaga2008adapting,
  title={Adapting robot behavior for human--robot interaction},
  author={Mitsunaga, Noriaki and Smith, Christian and Kanda, Takayuki and Ishiguro, Hiroshi and Hagita, Norihiro},
  journal={IEEE Transactions on Robotics},
  volume={24},
  number={4},
  pages={911--916},
  year={2008},
  publisher={IEEE}
}

@article{pirk2022protocol,
  title={A protocol for validating social navigation policies},
  author={Pirk, S{\"o}ren and Lee, Edward and Xiao, Xuesu and Takayama, Leila and Francis, Anthony and Toshev, Alexander},
  journal={arXiv preprint arXiv:2204.05443},
  year={2022}
}

@inproceedings{shiomi2010larger,
  title={A larger audience, please!—Encouraging people to listen to a guide robot},
  author={Shiomi, Masahiro and Kanda, Takayuki and Ishiguro, Hiroshi and Hagita, Norihiro},
  booktitle={2010 5th ACM/IEEE International Conference on Human-Robot Interaction (HRI)},
  pages={31--38},
  year={2010},
  organization={IEEE}
}

@inproceedings{palinko2016robot,
  title={Robot reading human gaze: Why eye tracking is better than head tracking for human-robot collaboration},
  author={Palinko, Oskar and Rea, Francesco and Sandini, Giulio and Sciutti, Alessandra},
  booktitle={2016 IEEE/RSJ International Conference on Intelligent Robots and Systems (IROS)},
  pages={5048--5054},
  year={2016},
  organization={IEEE}
}

@inproceedings{jensen2018knowing,
  title={Knowing you, seeing me: Investigating user preferences in drone-human acknowledgement},
  author={Jensen, Walther and Hansen, Simon and Knoche, Hendrik},
  booktitle={Proceedings of the 2018 CHI Conference on Human Factors in Computing Systems},
  pages={1--12},
  year={2018}
}

@inproceedings{he2016deep,
  title={Deep residual learning for image recognition},
  author={He, Kaiming and Zhang, Xiangyu and Ren, Shaoqing and Sun, Jian},
  booktitle={Proceedings of the IEEE conference on computer vision and pattern recognition},
  pages={770--778},
  year={2016}
}

@incollection{prechelt2002early,
  title={Early stopping-but when?},
  author={Prechelt, Lutz},
  booktitle={Neural Networks: Tricks of the trade},
  pages={55--69},
  year={2002},
  publisher={Springer}
}

@misc{candon2024react,
      title={REACT: Two Datasets for Analyzing Both Human Reactions and Evaluative Feedback to Robots Over Time}, 
      author={Kate Candon and Nicholas C. Georgiou and Helen Zhou and Sidney Richardson and Qiping Zhang and Brian Scassellati and Marynel Vázquez},
      year={2024},
      eprint={2402.00190},
      archivePrefix={arXiv},
      primaryClass={cs.RO}
}

@article{trautman2015robot,
  title={Robot navigation in dense human crowds: Statistical models and experimental studies of human--robot cooperation},
  author={Trautman, Pete and Ma, Jeremy and Murray, Richard M and Krause, Andreas},
  journal={The International Journal of Robotics Research},
  volume={34},
  number={3},
  pages={335--356},
  year={2015},
  publisher={SAGE Publications Sage UK: London, England}
}

@inproceedings{gockley2007natural,
  title={Natural person-following behavior for social robots},
  author={Gockley, Rachel and Forlizzi, Jodi and Simmons, Reid},
  booktitle={Proceedings of the ACM/IEEE international conference on Human-robot interaction},
  pages={17--24},
  year={2007}
}

@inproceedings{li2019comparing,
  title={Comparing human-robot proxemics between virtual reality and the real world},
  author={Li, Rui and van Almkerk, Marc and van Waveren, Sanne and Carter, Elizabeth and Leite, Iolanda},
  booktitle={2019 14th ACM/IEEE international conference on human-robot interaction (HRI)},
  pages={431--439},
  year={2019},
  organization={IEEE}
}

@article{karnan2022socially,
  title={Socially compliant navigation dataset (scand): A large-scale dataset of demonstrations for social navigation},
  author={Karnan, Haresh and Nair, Anirudh and Xiao, Xuesu and Warnell, Garrett and Pirk, S{\"o}ren and Toshev, Alexander and Hart, Justin and Biswas, Joydeep and Stone, Peter},
  journal={IEEE Robotics and Automation Letters},
  volume={7},
  number={4},
  pages={11807--11814},
  year={2022},
  publisher={IEEE}
}

@book{bartneck2020human,
  title={Human-robot interaction: An introduction},
  author={Bartneck, Christoph and Belpaeme, Tony and Eyssel, Friederike and Kanda, Takayuki and Keijsers, Merel and {\v{S}}abanovi{\'c}, Selma},
  year={2020},
  publisher={Cambridge University Press}
}

@article{martin2019jrdb,
  title={Jrdb: A dataset and benchmark for visual perception for navigation in human environments},
  author={Mart{\'\i}n-Mart{\'\i}n, Roberto and Rezatofighi, Hamid and Shenoi, Abhijeet and Patel, Mihir and Gwak, J and Dass, Nathan and Federman, Alan and Goebel, Patrick and Savarese, Silvio},
  journal={arXiv preprint arXiv:1910.11792},
  year={2019}
}

@article{breazeal2013crowdsourcing,
  title={Crowdsourcing human-robot interaction: New methods and system evaluation in a public environment},
  author={Breazeal, Cynthia and DePalma, Nick and Orkin, Jeff and Chernova, Sonia and Jung, Malte},
  journal={Journal of Human-Robot Interaction},
  volume={2},
  number={1},
  pages={82--111},
  year={2013},
  publisher={Journal of Human-Robot Interaction Steering Committee}
}

@article{toris2014robot,
  title={The robot management system: A framework for conducting human-robot interaction studies through crowdsourcing},
  author={Toris, Russell and Kent, David and Chernova, Sonia},
  journal={Journal of Human-Robot Interaction},
  volume={3},
  number={2},
  pages={25--49},
  year={2014},
  publisher={Citeseer}
}

@article{huck2021testing,
  title={Testing robot system safety by creating hazardous human worker behavior in simulation},
  author={Huck, Tom P and Ledermann, Christoph and Kr{\"o}ger, Torsten},
  journal={IEEE Robotics and Automation Letters},
  volume={7},
  number={2},
  pages={770--777},
  year={2021},
  publisher={IEEE}
}

@inproceedings{mitchell2020safety,
  title={Safety Perception and Behaviors during Human-Robot Interaction in Virtual Environments},
  author={Mitchell, Daxton and Choi, HeeSun and Haney, Justin M},
  booktitle={Proceedings of the Human Factors and Ergonomics Society Annual Meeting},
  volume={64},
  number={1},
  pages={2087--2091},
  year={2020},
  organization={SAGE Publications Sage CA: Los Angeles, CA}
}

@inproceedings{borgo2017crowdsourcing,
  title={Crowdsourcing for information visualization: Promises and pitfalls},
  author={Borgo, Rita and Lee, Bongshin and Bach, Benjamin and Fabrikant, Sara and Jianu, Radu and Kerren, Andreas and Kobourov, Stephen and McGee, Fintan and Micallef, Luana and von Landesberger, Tatiana and others},
  booktitle={Evaluation in the Crowd. Crowdsourcing and Human-Centered Experiments: Dagstuhl Seminar 15481, Dagstuhl Castle, Germany, November 22--27, 2015, Revised Contributions},
  pages={96--138},
  year={2017},
  organization={Springer}
}

@inproceedings{hwang2021ideabot,
  title={IdeaBot: investigating social facilitation in human-machine team creativity},
  author={Hwang, Angel Hsing-Chi and Won, Andrea Stevenson},
  booktitle={Proceedings of the 2021 CHI Conference on Human Factors in Computing Systems},
  pages={1--16},
  year={2021}
}

@article{inamura2021vr,
  title={VR platform enabling crowdsourcing of embodied HRI experiments--case study of online robot competition},
  author={Inamura, Tetsunari and Mizuchi, Yoshiaki and Yamada, Hiroki},
  journal={Advanced Robotics},
  volume={35},
  number={11},
  pages={697--703},
  year={2021},
  publisher={Taylor \& Francis}
}

@article{chetouani2021interactive,
  title={Interactive Robot Learning: An Overview},
  author={Chetouani, Mohamed},
  journal={ECCAI Advanced Course on Artificial Intelligence},
  pages={140--172},
  year={2021},
  publisher={Springer}
}

@inproceedings{knox2009interactively,
  title={Interactively shaping agents via human reinforcement: The TAMER framework},
  author={Knox, W Bradley and Stone, Peter},
  booktitle={Proceedings of the fifth international conference on Knowledge capture},
  pages={9--16},
  year={2009}
}

@inproceedings{macglashan2017interactive,
  title={Interactive learning from policy-dependent human feedback},
  author={MacGlashan, James and Ho, Mark K and Loftin, Robert and Peng, Bei and Wang, Guan and Roberts, David L and Taylor, Matthew E and Littman, Michael L},
  booktitle={International conference on machine learning},
  pages={2285--2294},
  year={2017},
  organization={PMLR}
}

@article{tian2021taxonomy,
  title={A taxonomy of social errors in human-robot interaction},
  author={Tian, Leimin and Oviatt, Sharon},
  journal={ACM Transactions on Human-Robot Interaction (THRI)},
  volume={10},
  number={2},
  pages={1--32},
  year={2021},
  publisher={ACM New York, NY, USA}
}

@article{che2020efficient,
  title={Efficient and trustworthy social navigation via explicit and implicit robot--human communication},
  author={Che, Yuhang and Okamura, Allison M and Sadigh, Dorsa},
  journal={IEEE Transactions on Robotics},
  volume={36},
  number={3},
  pages={692--707},
  year={2020},
  publisher={IEEE}
}

@inproceedings{zheng2019deephuman,
  title={Deephuman: 3d human reconstruction from a single image},
  author={Zheng, Zerong and Yu, Tao and Wei, Yixuan and Dai, Qionghai and Liu, Yebin},
  booktitle={Proceedings of the IEEE/CVF International Conference on Computer Vision},
  pages={7739--7749},
  year={2019}
}

@article{lew2023shutter,
  title={Shutter, the Robot Photographer: Leveraging Behavior Trees for Public, In-the-Wild Human-Robot Interactions},
  author={Lew, Alexander and Thompson, Sydney and Tsoi, Nathan and V{\'a}zquez, Marynel},
  journal={arXiv preprint arXiv:2302.00191},
  year={2023}
}

@article{collins2021review,
  title={A review of physics simulators for robotic applications},
  author={Collins, Jack and Chand, Shelvin and Vanderkop, Anthony and Howard, David},
  journal={IEEE Access},
  volume={9},
  pages={51416--51431},
  year={2021},
  publisher={IEEE}
}

@inproceedings{anderson2021sim,
  title={Sim-to-real transfer for vision-and-language navigation},
  author={Anderson, Peter and Shrivastava, Ayush and Truong, Joanne and Majumdar, Arjun and Parikh, Devi and Batra, Dhruv and Lee, Stefan},
  booktitle={Conference on Robot Learning},
  pages={671--681},
  year={2021},
  organization={PMLR}
}

@inproceedings{bharadhwaj2019data,
  title={A data-efficient framework for training and sim-to-real transfer of navigation policies},
  author={Bharadhwaj, Homanga and Wang, Zihan and Bengio, Yoshua and Paull, Liam},
  booktitle={2019 International Conference on Robotics and Automation (ICRA)},
  pages={782--788},
  year={2019},
  organization={IEEE}
}

@article{choi2021use,
  title={On the use of simulation in robotics: Opportunities, challenges, and suggestions for moving forward},
  author={Choi, HeeSun and Crump, Cindy and Duriez, Christian and Elmquist, Asher and Hager, Gregory and Han, David and Hearl, Frank and Hodgins, Jessica and Jain, Abhinandan and Leve, Frederick and others},
  journal={Proceedings of the National Academy of Sciences},
  volume={118},
  number={1},
  pages={e1907856118},
  year={2021},
  publisher={National Acad Sciences}
}

@inproceedings {HigginsISER2023,
   title	= {A Collaborative Building Task in VR vs. Reality},
   author	= {Padraig Higgins and Ryan Barron and Stephanie Lukin and Don Engel and Cynthia Matuszek},
   booktitle	= {Proc.~of the International Symposium on Experimental Robotics (ISER)},
   month	= {November},
   year		= {2023},
   location	= {Chiang Mai, Thailand}
}

@inproceedings{vazquez2020gaze,
  title={Gaze by Semi-Virtual Robotic Heads: Effects of Eye and Head Motion},
  author={V{\'a}zquez, Marynel and Milkessa, Yofti and Li, Michelle M and Govil, Neha},
  booktitle={2020 IEEE/RSJ International Conference on Intelligent Robots and Systems (IROS)},
  pages={11065--11071},
  year={2020},
  organization={IEEE}
}

@article{bai2022training,
  title={Training a helpful and harmless assistant with reinforcement learning from human feedback},
  author={Bai, Yuntao and Jones, Andy and Ndousse, Kamal and Askell, Amanda and Chen, Anna and DasSarma, Nova and Drain, Dawn and Fort, Stanislav and Ganguli, Deep and Henighan, Tom and others},
  journal={arXiv preprint arXiv:2204.05862},
  year={2022}
}

\end{document}